\documentclass[final]{colt2024} 

\usepackage{times}
\usepackage{mymath, natbib, enumitem, xfrac,xcolor}
\usepackage{booktabs,makecell}

\usepackage[normalem]{ulem} 

\usepackage[noabbrev]{cleveref}
\crefname{equation}{}{}
\crefname{assumption}{assumption}{assumptions}
\newlist{assumpenum}{enumerate}{1} 
\setlist[assumpenum]{label=\Alph*., ref=\theassumption\Alph*, leftmargin=*}
\crefalias{assumpenumi}{assumption}

\newenvironment{proofof}[1][]{\begin{proof}\textbf{(of {#1})~}}{\end{proof}}

\def\showcomments{}
\ifdefined\showcomments
  \newcommand\JW[1]{\textcolor{red}{[JW: #1]}}
  \newcommand\MT[1]{\textcolor{green!50!black}{[MJT: #1]}}
  \newcommand\PLB[1]{\textcolor{cyan}{[PB: #1]}}
\else
  \newcommand\JW[1]{}
  \newcommand\MT[1]{}
  \newcommand\PLB[1]{}
  \newcommand\sketchy[1]{}
\fi

\title[Large Stepsize Gradient Descent for Logistic Loss]{Large Stepsize Gradient Descent for Logistic Loss:\\ Non-Monotonicity of the Loss Improves Optimization Efficiency}

\coltauthor{%
 \Name{Jingfeng Wu} \Email{uuujf@berkeley.edu}\\
 \addr University of California, Berkeley
 \AND
 \Name{Peter L. Bartlett}\thanks{Alphabetical order} \Email{peter@berkeley.edu}\\
 \addr University of California, Berkeley and Google DeepMind%
 \AND
 \Name{Matus Telgarsky}\footnotemark[1] \Email{mjt10041@nyu.edu}\\
 \addr New York University%
 \AND
 \Name{Bin Yu}\footnotemark[1] \Email{binyu@berkeley.edu}\\
 \addr University of California, Berkeley%
}

\begin{document}
\maketitle

\allowdisplaybreaks

\begin{abstract}
We consider \emph{gradient descent} (GD) with a constant stepsize applied to logistic regression with linearly separable data, where the constant stepsize $\eta$ is so large that the loss initially oscillates. We show that GD exits this initial oscillatory phase rapidly --- in $\Ocal(\eta)$ steps, and subsequently achieves an $\tilde{\Ocal}(1 / (\eta t) )$ convergence rate after $t$ additional steps. Our results imply that, given a budget of $T$ steps, GD can achieve an \emph{accelerated} loss of $\tilde{\Ocal}(1/T^2)$ with an aggressive stepsize $\eta:= \Theta( T)$, without any use of momentum or variable stepsize schedulers. Our proof technique is versatile and also handles general classification loss functions (where exponential tails are needed for the $\tilde{\Ocal}(1/T^2)$ acceleration), nonlinear predictors in the \emph{neural tangent kernel} regime, and online \emph{stochastic gradient descent} (SGD) with a large stepsize, under suitable separability conditions. 
\end{abstract}


\section{Introduction}
Gradient-based methods are arguably the most popular optimization methods in modern machine learning. 
The simplest version, constant-stepsize \emph{gradient descent} (GD), is defined as
\begin{equation}\label{eq:gd}\tag{GD}
    \wB_{t} := \wB_{t-1} - \eta \grad L(\wB_{t-1}), \quad t= 1,2,\dots, T,
\end{equation}
where $\wB \in \Rbb^{d}$ is the trainable parameter, $L(\cdot)$ is the loss function, $\eta>0$ is a stepsize fixed across the iterates, and $\wB_0$ is the initialization.
We focus on understanding GD with a \emph{large} stepsize, which means that the induced loss 
does not decrease monotonically. 

The classical theory for constant-stepsize GD assumes a small stepsize such that the induced loss decreases monotonically \citep{nesterov2018lectures}. 
For example, if $L(\wB)$ is $\beta$-smooth, then choosing $\eta < 2 /\beta$ guarantees $L(\wB_t)$ to decrease monotonically, a result known as the \emph{descent lemma} \citep[Section 1.2.3]{nesterov2018lectures}. 
Fruitful convergence results have been developed for small stepsize GD based on the descent lemma or its variants. 
However, little theory is known when GD is run with a large stepsize (for instance, $\eta > 2/\beta$ for a $\beta$-smooth function), which induces non-monotonic loss values (exceptions will be discussed later in \Cref{sec:related}). 
\citet{cohen2020gradient} refer to this as the \emph{edge of stability} (EoS) regime, and they further point out that GD usually operates in the EoS regime for trained neural networks to exhibit reasonable optimization or generalization performance \citep{wu2018sgd,cohen2020gradient}.

\begin{figure}[t]
 \centering
\subfigure[Training loss, toy dataset.]{
\label{fig:4data:risk}
\includegraphics[width=0.45\linewidth]{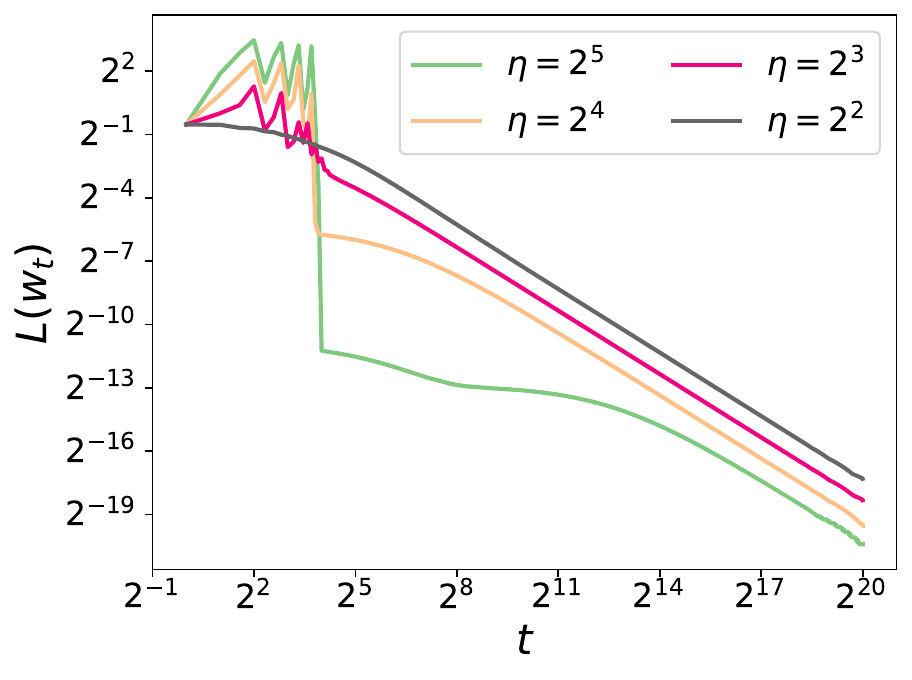}
}
\hfill
\subfigure[Asymptotic rate, toy dataset.]{
\label{fig:4data:rate}
\includegraphics[width=0.45\linewidth]{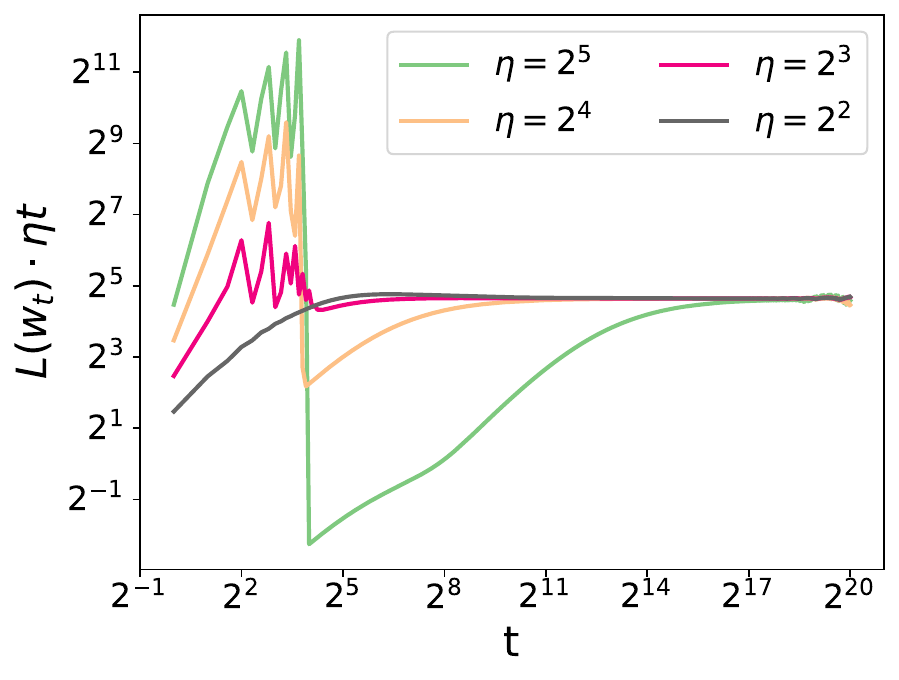}
}
\subfigure[Training loss, subset of MNIST.]{
\label{fig:mnist:risk}
\includegraphics[width=0.45\linewidth]{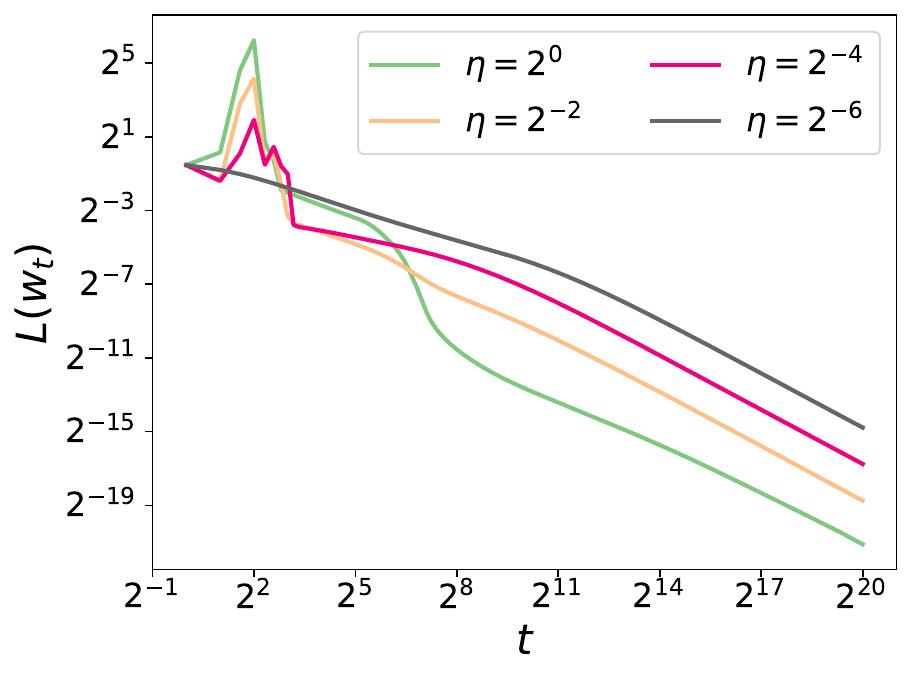}
}
\hfill
\subfigure[Asymptotic rate, subset of MNIST.]{
\label{fig:mnist:rate}
\includegraphics[width=0.45\linewidth]{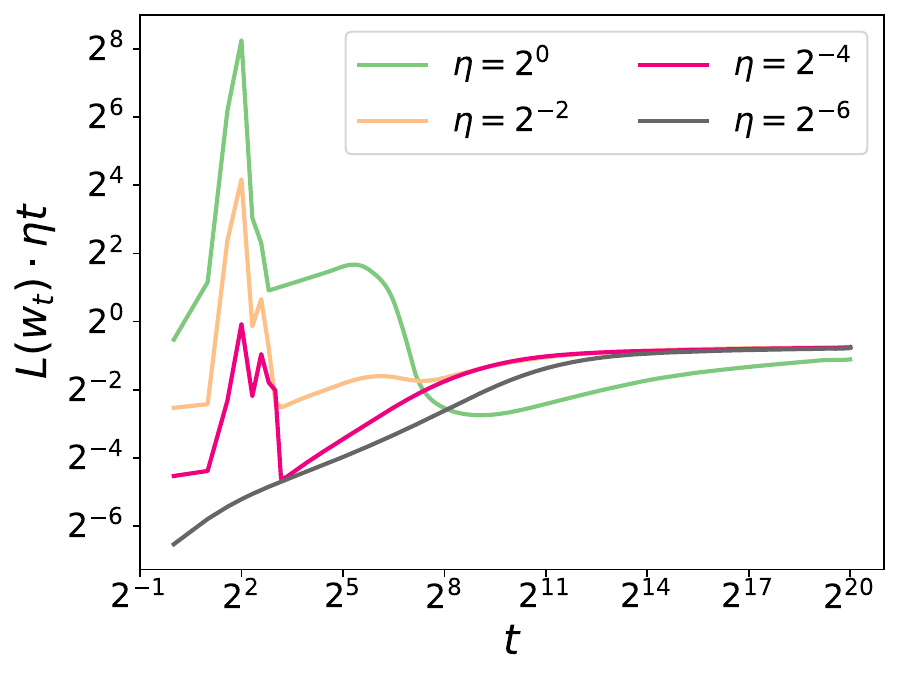}
}
\caption{
Behaviors of \eqref{eq:gd} for logistic regression \eqref{eq:logistic-regression} with initialization $\wB_0:=0$.
\Cref{fig:4data:risk,fig:4data:rate} are results on a toy dataset consisting of four samples, $\xB_1 = (1,0.2),\ y_1=1$, $\xB_2=(-2,0.2),\ y_2=1$, $\xB_3 = (-1,-0.2),\ y_3=-1$, and $\xB_4 = (2, -0.2),\ y_4=-1$.
Here, the stepsize $\eta=2^{2}$ is small, and $\eta \in \{2^{3}, 2^{4}, 2^{5}\}$ is large (depending on whether the loss decreases monotonically).
\Cref{fig:mnist:risk,fig:mnist:rate} are results on $1000$ samples from the MNIST dataset with labels ``0'' or ``8''.
Here, the stepsize $\eta=2^{-6}$ is small, 
and $\eta \in \{2^{-4},2^{-2}, 2^{0}\}$ is large.
\Cref{fig:4data:risk,fig:mnist:risk} suggest that \eqref{eq:gd} with a larger stepsize incurs a smaller loss \emph{when stabilized}. 
\Cref{fig:4data:rate,fig:mnist:rate} suggest that  $L(\wB_t) = \Theta( 1/(\eta t))$ asymptotically as $t\to \infty$.
}
\label{fig:mnist}
\end{figure}

This work considers large stepsize GD in minimal yet natural machine learning settings, such as \emph{logistic regression}:
\begin{equation}\label{eq:logistic-regression}
    L(\wB):= \frac{1}{n} \sum_{i=1}^n \ln \big(1+\exp(-y_i \xB_i^\top \wB ) \big),\quad 
    (\xB_i, y_i)\in \Rbb^{d}\times\{\pm 1\}.
\end{equation}
Our first main result (\Cref{thm:logistic-reg}) characterizes the dynamics of large stepsize GD  for logistic regression with \emph{linearly separable} data. 
We show that GD initially induces a non-monotonic loss (the EoS phase) and then exits this regime in finite time (phase transition), and afterwards the loss decreases monotonically (the stable phase).
Specifically:
\begin{enumerate}[leftmargin=*]
\item \textbf{The EoS phase.}~  
First, we show the loss \emph{averaged} over $t$ steps decreases at a rate of $\tilde\Ocal\big((1+\eta^2) / (\eta t)\big)$ for GD with an arbitrarily large stepsize $\eta$. 
In particular, this applies to GD in the EoS phase where the loss oscillates locally.

\item \textbf{Phase transition.}~
Second, we show that GD exits the initial EoS phase (if it ever enters) in $\Ocal(\eta)$ steps. Then GD undergoes a phase transition and 
enters a stable phase,  where the loss decreases monotonically, and stays in this phase.  

\item \textbf{The Stable phase.}~
Finally, we show in the stable phase, the loss decreases monotonically at an $\tilde\Ocal\big(1/ (\eta t)\big)$ rate after $t$ steps. 
\Cref{fig:4data:rate,fig:mnist:rate} empirically verify that this rate is sharp asymptotically (as $t\to \infty$) ignoring logarithmic factors. 
\end{enumerate}
The above result immediately justifies the \textbf{benefits of a large stepsize} for improving optimization efficiency.
First, a larger $\eta$ yields a smaller constant factor in the \emph{asymptotic} bound, $\tilde\Ocal(1/(\eta t))$.
We highlight that $\eta$ can be an arbitrarily large constant, beyond the classical wisdom that constrains $\eta$ by the inverse smoothness \citep{nesterov2018lectures}. 
Second, given a budget of $T$ steps, GD attains an $\tilde\Ocal(1/T^2)$ loss when equipped with an aggressive stepsize $\eta:= \Theta(T)$ that optimally balances the steps spent in the EoS and stable phases (see \Cref{cor:acceleration}).
In contrast, we also provide an $\Omega(1/T)$ loss lower bound for constant-stepsize GD that does not enter the EoS phase (see \Cref{thm:gd:lb}). 
Our theory explains the empirical observations in \Cref{fig:mnist}.

We extend the above result in three notable aspects.
\begin{itemize}[leftmargin=*]
\item \textbf{General losses.}~
We relax the logistic loss in \eqref{eq:logistic-regression} to general classification loss functions under a set of conditions (see \Cref{assump:loss} and \Cref{thm:ntk}). 
Specifically, the EoS bound relies on the \emph{Lipschitzness} of the loss, the stable phase bound relies on an additional \emph{self-boundedness} condition, and the $\tilde\Ocal(1/T^2)$ acceleration relies on the loss having an \emph{exponential tail}.

\item \textbf{Nonlinear models.}~
We extend the linear predictor in \eqref{eq:logistic-regression} to a two-layer network in the \emph{neural tangent kernel} \citep[NTK,][]{jacot2018neural}
regime (\Cref{thm:ntk}). 
To our knowledge, this is the first NTK result that accommodates an arbitrarily large stepsize (the minimum network width depends on the stepsize). 
For a constant stepsize, the minimum network width is \emph{polylogarithmic} in the number of steps and the number of samples, recovering the main result of \citet{ji2019polylogarithmic} as a special case.

\item \textbf{SGD.}~
Finally, we consider constant-stepsize online \emph{stochastic gradient descent} (SGD) for logistic regression with separable data.
We provide upper bounds on the population logistic loss and the population zero-one error (\Cref{thm:sgd:linear}), allowing the stepsize to be $o(T)$ (where $T$ is the maximum number of steps) and arbitrarily large, respectively. 
These are in contrast to the typical SGD results in online smooth convex optimization that only allow small (even vanishing) stepsizes \citep[see, for example,][Chapter 3]{hazan2022introduction}. 
\end{itemize}

\paragraph{Implication and limitations of our theory in practice.}
The work by \citet{cohen2020gradient} empirically identifies two phenomena when training neural networks using GD with a large stepsize~($\eta$): \emph{progressive sharpening}, where the sharpness increases until around $2/\eta$, and EoS, where the sharpness oscillates around $2/\eta$.
Their definition of EoS and ours are equivalent for quadratic functions, and our definition highlights the unstable loss behaviors more explicitly for non-quadratic functions. 
Additionally, a transition from the EoS phase to the stable phase is observed in neural networks trained with cross-entropy loss \citep[see][Appendix A]{cohen2020gradient}. 
Our theory offers valuable insights into the EoS and stable phases but remains inconclusive regarding the progressive sharpening phenomenon.

\subsection{Related Works}\label{sec:related}

\paragraph{Large stepsize and EoS.}
Empirically, training neural networks with GD (or large batch SGD) often requires a large stepsize to achieve reasonable optimization and generalization performance \citep[see, for example,][]{hoffer2017train,goyal2017accurate,wu2018sgd,cohen2020gradient}, whereby the training loss exhibits non-monotonic behavior.
\citet{cohen2020gradient} introduce the term \emph{edge of stability} (EoS) to refer to this.
The theory of EoS and large stepsize has been studied in various cases, such as one- or two-dimensional (nearly analytic) functions \citep{zhu2022understanding,ahn2023learning,kreisler2023gradient,chen2023stability,wang2023good}, scale-invariant networks \citep{lyu2022understanding}, diagonal linear networks \citep{even2023s,andriushchenko2023sgd}, and matrix factorization \citep{wang2022large,chen2023beyond}.
Besides, there are some general theories for EoS \citep{kong2020stochasticity,ahn2022understanding,damian2022self,ma2022beyond,wang2022analyzing,lu2023benign}, but they are built up on subtle assumptions. 
In comparison, we work in a natural classification setting under standard assumptions.

Among the literature of EoS, the work by \citet{wu2023implicit} is most relevant to us. 
They show the convergence of GD with an arbitrarily large stepsize for logistic regression with linearly separable and \emph{non-degenerate} data. 
Our work significantly improves theirs. 
First, the constant factors in their bounds are \emph{exponential} in the stepsize $\eta$.
In comparison, we nail down the dependence on $\eta$ in each phase (see \Cref{thm:logistic-reg}), establishing the $\tilde\Ocal(1/(\eta t))$ asymptotic rate and the $\tilde\Ocal(1/T^2)$ acceleration with stepsize $\eta = \Theta(T)$ (see \Cref{cor:acceleration}).  
So we explain the benefits of large stepsizes while they cannot. 
Second, they require the dataset to be \emph{non-degenerate} such that the support vectors span the dataset (see their Assumption 3), while we do not need this condition. 
Finally, their results are specialized to GD and logistic regression, while our approach is versatile and also handles general classification losses, nonlinear predictors in the NTK regime, and SGD with a large stepsize.   

\paragraph{Variable stepsize schedulers.}
AdaBoost \citep{freund1997decision}, in the language of convex optimization, is \emph{coordinate
descent} with an exponentially increasing stepsize scheduler, achieving an $\exp(-\Omega(t))$ rate \citep[see][for example]{telgarsky2013margins}. 
Following this idea, a line of work \citep{nacson2019convergence,ji2021characterizing,ji2021fast} considers GD with an increasing stepsize scheduler (that adapts to the loss value) and obtains faster rates for linear classification under exponentially tailed losses.
Despite the increase in stepsize, the loss in these works \emph{always} decreases monotonically as the gradient dynamics satisfy a local version of the descent lemma. 
In contrast, we focus on a large stepsize that allows GD to enter the EoS phase where the loss oscillates. 
We remark that our results for constant stepsize can be applied recursively for analyzing variable stepsize schedulers, but this is beyond the scope of the current paper.

A recent line of work uses variable stepsize schedulers to accelerate the \emph{worst-case} rate of GD for a class of problems \citep[][and references therein]{kelner2022big,altschuler2023acceleration-convex}. 
In particular, \citet{altschuler2023acceleration-convex} show that GD with the \emph{Silver} stepsize scheduler achieves an $\Ocal(1 / t^{1.2716})$ rate for smooth convex optimization. 
Similarly to our work, they achieve acceleration by using aggressive stepsizes that result in oscillatory losses. 
However, they design a delicate variable stepsize scheduler that improves the worst-case rate of GD, while we focus on constant-stepsize GD for 
linear classification problems. 

\citet{malitsky2020adaptive} design an adaptive stepsize scheduler for GD for smooth convex optimization, where they show a convergence rate without an explicit dependence on the global smoothness parameter. Their analysis involves two consecutive GD steps, instead of using the classical descent lemma. Thus their analysis also allows non-monotonic loss. In comparison, we focus on GD with a large but constant stepsize, and our analysis relies on the separability condition of the dataset instead of variable stepsizes. 

\paragraph{Neural tangent kernel.} 
The \emph{neural tangent kernel} (NTK) is most prominently presented and named by \citet{jacot2018neural},
and similar concepts are used by 
subsequent works to analyze the early training dynamics of GD.
In the majority of these works \citep[see][for example]{du2018gradient,zou2018stochastic,zou2019improved,allen2019convergence}, the stepsize of GD is $o(1)$ and vanishes as a function of other parameters such as the target error or the level of overparameterization.
The only two exceptions are by  \citet{ji2019polylogarithmic} and \citet{chen2020much}, allowing a constant stepsize smaller than $2/\beta$, where $\beta$ is the smoothness of the objective near initialization.
In comparison, our NTK result (see \Cref{thm:ntk}) handles arbitrarily large stepsizes, in particular, it recovers the related result of \citet{ji2019polylogarithmic} when specializing the stepsize to a constant.


\section{GD for Logistic Regression}\label{sec:logistic-reg}
We now present our results on GD for logistic regression.
We assume the dataset is bounded and linearly separable.

\begin{assumption}[Bounded and separable data]\label{assump:bounded-separable}
Assume the training data $(\xB_i, y_i)_{i=1}^n$ satisfies
\begin{assumpenum}
\item for every $i=1,\dots,n$,
\(
 \| \xB_i \| \le 1
\)
and
 $y_i \in \{\pm 1\}$;
\item 
there is a {\em margin} $\gamma>0$ and a unit vector $\wB_*$ such that
\(
\la y_i \xB_i, \, \wB_* \ra \ge \gamma 
\)
for every $i=1,\dots,n$.
\end{assumpenum}
\end{assumption}

The next theorem characterizes the behaviors of GD for logistic regression.
\begin{theorem}[Bounds on risk and phase transition time]\label{thm:logistic-reg}
Consider \cref{eq:gd} with stepsize $\eta>0$ for logistic regression \cref{eq:logistic-regression} under \Cref{assump:bounded-separable}.
Without loss of generality, assume $\wB_0 = 0$. We say that GD is {\em in the stable phase} at step $t$ when $L(w_t)$ decreases monotonically from $t$ onwards, and {\em in the EoS phase} when it does not. Then we have the following:
\begin{itemize}[leftmargin=*]
    \item \textbf{The EoS phase.} 
    For every $t > 0$ (and in particular in the EoS phase), we have 
\begin{align*}
    \frac{1}{t}\sum_{k=0}^{t-1} L(\wB_k) \le \frac{1 + \ln^2(\gamma^2 \eta t) + \eta^2/4}{ \gamma^2 \eta t}.
\end{align*}
\item \textbf{The stable phase.} 
If $s$ is such that 
\begin{equation}\label{eq:stable-phase-criteria}
    L(\wB_s) \le \frac{1}{ \eta},
\end{equation}
then \eqref{eq:gd} is in the stable phase, that is, 
$\big(L(\wB_t)\big)_{t\ge s}$ decreases monotonically, and moreover,
\begin{align*}
    L(\wB_t) \le \frac{2F(\wB_s) + \ln^2 (\gamma^2 \eta (t-s)) }{\gamma^2 \eta (t-s)},\quad \text{where}\ \ F(\wB) := \frac{1}{n}\sum_{i=1}^n \exp( -y_i \xB_i^\top \wB).
\end{align*}
\item \textbf{Phase transition time.}
There exists $s\le \tau$ such that \cref{eq:stable-phase-criteria} holds and \(F(\wB_s) \le 1\), where
\begin{equation*}
    \tau := \frac{60}{\gamma^2}\max\bigg\{\eta,\ n,\  e,\ \frac{\eta + n}{\eta}\ln\frac{\eta + n}{\eta} \bigg\}.
\end{equation*}
\end{itemize}
\end{theorem}

The proof of \Cref{thm:logistic-reg} is deferred to \Cref{append:sec:logistic-proof}.
\Cref{thm:logistic-reg} provides control over the average loss in the EoS phase, the last iterate loss in the stable phase, and the maximum length of the EoS phase.
Note that GD might never enter the EoS phase. For instance, this happens when the stepsize is sufficiently small such that \cref{eq:stable-phase-criteria} holds for $\wB_s = \wB_0 = 0$.
However, if GD enters the EoS phase at the beginning of the optimization, the algorithm must undergo a phase transition and switch to the stable phase after a finite time.
We emphasize that \Cref{thm:logistic-reg} allows an arbitrarily large $\eta$.

\paragraph{Benefits of large stepsizes.}
\Cref{thm:logistic-reg} shows that a large stepsize improves optimization efficiency in two ways. 
First, when $\eta = \Ocal(1)$, the EoS phase ends in finite time, so asymptotically GD stays in the stable phase and attains an $\tilde\Ocal\big(1/(\eta t)\big)$ rate. 
In this case, a larger stepsize leads to a smaller constant factor.
We also note that this rate can be sharp ignoring logarithmic factors according to the experimental results in \Cref{fig:4data:rate,fig:mnist:rate}.
Interestingly, recall the classical rate of GD for smooth convex optimization is $\Ocal\big(1/(\eta t)\big)$ for $\eta < 2/\beta$, where $\beta$ is the smoothness parameter \citep[][Theorem 2.1.14]{nesterov2018lectures}. 
Our results (nearly) match this rate but remove the $\eta < 2/\beta$ condition for logistic regression, a special class of smooth convex optimization problems.

Second, \Cref{thm:logistic-reg} suggests that GD with a larger stepsize converges faster in the stable phase, but it takes more steps to exit the initial EoS phase. 
Picking a stepsize proportional to the total number of steps will balance these two effects and lead to an acceleration effect. 
This is formalized by the following corollary 
(the proof is deferred to \Cref{append:sec:acceleration}).


\begin{corollary}[Acceleration of a large stepsize]\label[corollary]{cor:acceleration}
Under the same setup as in \Cref{thm:logistic-reg}, 
consider \cref{eq:gd} with a given budget of $T$ steps, where
\(
T \ge 120 \max\{e,n\} / \gamma^2.
\)
Then for 
\[
\eta := \frac{\gamma^2}{120} T,
\]
we have $\tau \le T/2$ and
\[
L(\wB_T) \le   480\frac{\ln^2(\gamma^4 T^2)}{\gamma^4 T^2} = \Ocal\bigg(\frac{\ln^2(T)}{T^2}\bigg).
\]
\end{corollary}

\Cref{cor:acceleration} shows that GD attains an $\tilde\Ocal(1/T^2)$ loss after $T$ steps with a large stepsize $\eta = \Theta(T)$.
We point out that the final low loss is a consequence of the initial EoS phase. 
In particular, 
the next theorem (the proof is deferred to \Cref{append:sec:lower-bound}) shows that in general, GD with $T$ steps of fixed stepsize must suffer an $\Omega(1/T)$ loss if it never enters the EoS phase. 

\begin{theorem}[Lower bound in the classical regime]\label[theorem]{thm:gd:lb}
Consider \cref{eq:gd} with $\wB_0=0$ and stepsize $\eta>0$ for logistic regression \cref{eq:logistic-regression} on the following dataset:
\begin{align*}
    \xB_1 = \big( \gamma,\, \sqrt{1-\gamma^2} \big), \quad \xB_2 = \big( \gamma,\, - \sqrt{1-\gamma^2}/2 \big),\quad y_1 =y_2= 1,\quad 0<\gamma<0.1.
\end{align*}
It is clear that $(\xB_i, y_i)_{i=1,2}$ satisfy \Cref{assump:bounded-separable}.
If $\big( L(\wB_t) \big)_{t\ge 0}$ is non-increasing, 
then
\[L(\wB_t) \ge c_0 / t,\quad t\ge 1,\]
where $c_0>0$ is a function of $\gamma$ but is independent of $t$ and $\eta$.
\end{theorem}

We remark that the acceleration in \Cref{cor:acceleration} requires the budget to be at least linear in the sample size, that is, $T\ge \Omega(n)$. This requirement is rooted in the phase transition time bound in \Cref{thm:logistic-reg}. We conjecture our phase transition time bound, especially its dependence on the sample size, is improvable, and thus the $T\ge \Omega(n)$ condition in \Cref{cor:acceleration} can be relaxed. Some empirical evidence is provided in \Cref{fig:mnist-acc} in \Cref{append:sec:additional-exp}.

\paragraph{A proof technique.}
We now showcase a powerful yet surprisingly simple technique for handling large stepsizes, which lies at the heart of the proof of \Cref{thm:logistic-reg}.
To begin with, consider a comparator 
\(\uB := \uB_1  + \uB_2\). Then from the definition of \eqref{eq:gd}, we have
\begin{align*}
    \|\wB_{t+1} - \uB \|^2 
    &= \| \wB_t-\uB\|^2 + 2\eta \la \grad L(\wB_t) , \uB -\wB_t \ra + \eta^2\big\|\grad L(\wB_t) \big\|^2_2 \\
     &= \| \wB_t-\uB\|^2 + 2\eta \la \grad L(\wB_t) , \uB_1-\wB_t \ra  + \eta^2 \Big( \la \grad L(\wB_t) ,  2\uB_2/\eta \ra + \big\|\grad L(\wB_t) \big\|^2_2 \Big).
\end{align*} 
The last term is \emph{non-positive} when we choose $\uB_2 := \wB_* \cdot \eta / (2\gamma) $, because 
\begin{equation*}
    \la \grad L(\wB_t) ,  2\uB_2/\eta \ra
    = \frac{1}{n}\sum_{i=1}^n \ell'(y_i\xB_i^\top \wB_t )\la y_i \xB_i, 2\uB_2/\eta \ra
    \le \frac{1}{n}\sum_{i=1}^n \ell'(y_i\xB_i^\top \wB_t),
\end{equation*}
where we use $\la y_i\xB_i, \wB_*\ra \ge \gamma$ and $\ell'(\cdot) \le 0$,
and 
\begin{equation*}
  \big\|\grad L(\wB_t) \big\|^2_2 = \bigg\|\frac{1}{n} \sum_{i=1}^n \ell'(y_i \xB_i^\top \wB_t ) y_i \xB_i \bigg\|^2_2 
  \le \bigg(\frac{1}{n}\sum_{i=1}^n \ell'(y_i \xB_i^\top \wB_t )\bigg)^2
  \le \frac{1}{n}\sum_{i=1}^n |\ell'(y_i \xB_i^\top \wB_t )|,
\end{equation*}
where we use $\|y_i\xB_i\|\le 1$ and $|\ell'(\cdot)| \le 1$.
Dropping the last term, we have
\begin{equation*}
    \|\wB_{t+1} - \uB \|^2 
    \le \| \wB_t-\uB\|^2 + 2\eta \la \grad L(\wB_t),  \uB_1- \wB_t \ra  \le  \| \wB_t-\uB\|^2 + 2\eta \big( L(\uB_1) - L( \wB_t) \big),
\end{equation*} 
where we use the convexity of $L(\cdot)$.
Telescoping the sum from $0$ to $t$ and rearranging, we obtain
\begin{equation}\label{eq:risk-iter-bound}
   \frac{  \|\wB_t - \uB \|^2}{2\eta t} + \frac{1}{t}\sum_{k=0}^{t-1} L(\wB_k) 
   \le  L(\uB_1) + \frac{ \| \wB_0 - \uB\|^2 }{2\eta t}.
\end{equation}
Let us choose $\uB_1 := \wB_* \cdot \ln(\gamma^2 \eta t) / \gamma$, then we have $L(\uB_1) \le 1/(\gamma^2 \eta t)$ and $\|\wB_0-\uB\| = \|\uB\| = \ln(\gamma^2 \eta t)/\gamma  + \eta / (2\gamma)$.
Then \eqref{eq:risk-iter-bound} allows us to control the average loss and the norm of $\wB_t$ under any stepsize $\eta>0$.
We refer the reader to \Cref{append:sec:logistic-proof} for a complete proof of \Cref{thm:logistic-reg}.

\paragraph{Other acceleration techniques.}
Our focus in this paper is to understand the benefits of using a large but constant stepsize in GD. Based on a simple yet powerful technique, we show that GD with a suitably large stepsize leads to non-monotonic loss and achieves improved optimization efficiency. 
Our techniques are orthogonal to standard acceleration methods such as momentum and variable stepsize schedulers. 
We conjecture these techniques can be combined to obtain an even faster optimization. This is left for future work.

\paragraph{Implicit bias and generalization.}
We conclude this part by discussing the implicit bias and generalization of the output of GD with a large stepsize, $\eta$. 
When $\eta = \Ocal(1)$ compared to the number of steps, we can apply existing implicit bias results \citep{soudry2018implicit,ji2018risk} to GD in the stable phase (replacing the initialization by $\wB_s$) to show the direction of the GD iterate asymptotically converges to the max-margin direction.
In this situation, the generalization of the GD output is guaranteed by standard margin-based generalization bounds \citep[][for example]{bartlett1999generalization}. 

When $\eta\ge \omega(1)$ compared to the total number of steps $T$ (for example, $\eta=\Theta(T)$ as in \Cref{cor:acceleration}), existing implicit bias results \citep{soudry2018implicit,ji2018risk} can no longer be applied. 
In particular, our bound on the norm of the GD iterates is as large as $\Ocal(\eta)$ (see \eqref{eq:risk-iter-bound} or \Cref{lemma:parameter-risk-bound} in \Cref{append:sec:logistic-proof}), which prevents us from proving a good margin for the GD iterates. 
Nonetheless, as the final GD iterate attains low training error, the standard uniform-convergence theory is applicable to guarantee its generalization. We refer the reader to \Cref{prop:gd:linear:generalization} in \Cref{append:sec:vc-dim} for one such bound.

We leave for future work the study of the implicit bias and generalization of GD with a very large stepsize, that is, $\eta\ge \omega(1)$ compared to the total number of steps $T$.

\section{SGD for Logistic Regression}
We next consider constant-stepsize online \emph{stochastic gradient descent} (SGD) for logistic regression, defined as
\begin{equation}\label{eq:sgd}\tag{SGD}
    \wB_{t+1} := \wB_{t} - \eta \grad L_t(\wB_{t}),\ \  \text{where} \ \ L_t(\wB):= \ln\big(1 + \exp(- y_t \xB_t^\top \wB ) \big),\quad t\ge 0.
\end{equation}
Here, $(\xB_t, y_t)_{t\ge 0}$ are independent and identically distributed according to the following assumption.
\begin{assumption}[Bounded and separable distribution]\label{assump:sgd:bounded-separable}
Assume that $(\xB_t, y_t)_{t\ge 0}$ are independent copies of $(\xB, y)$ that follows a distribution such that
\begin{assumpenum}
\item the label is binary, $y \in \{\pm 1\}$, and
\(
\| \xB \| \le 1,
\)
almost surely;
\item there exist a margin $\gamma>0$ and a unit vector $\wB_*$ such that
\(
\la y \xB, \wB_* \ra \ge \gamma,
\)
almost surely.
\end{assumpenum}
\end{assumption}

The following theorem controls the population logistic loss and the population zero-one loss of SGD with a large stepsize.
\begin{theorem}[Risk and error bounds]\label{thm:sgd:linear}
Consider \cref{eq:sgd} with stepsize $\eta > 0$ for logistic regression under \Cref{assump:sgd:bounded-separable}.
Without loss of generality, assume $\wB_0 =0$.
Denote 
\[
L(\wB) := \Ebb \ln\big(1 + \exp(- y \xB^\top \wB ) \big),
\]
where the expectation is over $(\xB, y)$ as in \Cref{assump:sgd:bounded-separable}.
Then with probability at least $1-\delta$ over the randomness of $(\xB_k, y_k)_{k=0}^{t-1}$, we have 
\begin{align*}
\frac{1}{t}  \sum_{k=0}^{t-1} L(\wB_k) 
\le \frac{2+2\ln^2 (\gamma^2 \eta t) + \eta^2 /2}{\gamma^2 \eta  t} + \frac{3 + 2\ln(\gamma^2 \eta t) + \eta }{\gamma} \cdot \frac{18\ln(1/\delta)}{t},
\end{align*}
and
\begin{align*}
   \frac{1}{t}  \sum_{k=0}^{t-1} \Pr(y \xB^\top \wB_k \le 0 ) 
   \le \frac{4\big(\sqrt{2} +2\ln(\gamma^2 \eta t) + \eta \big)}{\gamma^2 \eta t} + \frac{36\ln(1/\delta)}{t}.
\end{align*}
\end{theorem}

The proof of \Cref{thm:sgd:linear} is deferred to \Cref{append:sec:sgd-proof}.
The population logistic loss bound in \Cref{thm:sgd:linear} allows SGD with a stepsize as large as $o(t)$. For instance, in $t$ steps, SGD achieves an $\Ocal(1/\sqrt{t})$ population logistic loss on average with $\eta := \sqrt{t}$.
This is in contrast to the typical online smooth convex optimization results, where the stepsize for SGD is a small constant or even vanishing \citep[see][Chapter 3, for example]{hazan2022introduction}. 

The population zero-one error bound in \Cref{thm:sgd:linear} allows SGD with an arbitrarily large stepsize. 
In particular, for a large stepsize $\eta \ge \ln(\gamma^2 t)$, there exists $\hat\wB\in (\wB_k)_{k=0}^{t-1}$ such that
\[
\Pr\big(y \xB^\top \hat\wB \le 0 \big) \le C \bigg( \frac{1}{\gamma^2 t} + \frac{\ln(1/\delta)}{t} \bigg),
\]
where $C\ge 1$ is an absolute constant. 
This bound for SGD matches that achieved by the Perceptron algorithm \citep{novikoff1962convergence,hanneke2021stable}, and matches (or improves by logarithmic factors) that of large margin classifiers \citep{bartlett1999generalization,gronlund2020near,hanneke2021stable}.
Note that the parameter $\hat\wB$ might not exhibit a large margin over the training data. 

Unlike the result in \Cref{sec:logistic-reg}
for batch GD, \Cref{thm:sgd:linear} does not show the benefits of a large stepsize for SGD (except on logarithmic factors). 
Specifically, 
the logistic loss bound in \Cref{thm:sgd:linear} is minimized when $\eta = \Theta\big(\ln(t)\big)$ (and larger stepsize such as $\eta \ge \omega\big(\ln(t)\big)$ hurts the bound), and the zero-one error bound in \Cref{thm:sgd:linear} is minimized when $\eta \ge \Omega \big(\ln(t)\big)$ (so larger stepsize is as good).
However, both optimized bounds do not improve the bounds for $\eta = \Theta(1)$ ignoring logarithmic factors.
We leave it as an open problem to investigate whether or not a large stepsize benefits SGD. 


Finally, the above discussion applies to multi-pass SGD, empirical logistic loss, and empirical zero-one error by setting the distribution in \Cref{assump:sgd:bounded-separable} to an empirical distribution.

\section{General Loss Functions and Nonlinear Predictors}
We now extend our results in \Cref{sec:logistic-reg} to general classification loss functions and a two-layer network in the \emph{neural tangent kernel} \citep[NTK,][]{jacot2018neural} regime.

\paragraph{General classification losses.}
We first introduce a set of conditions for the loss functions.
\begin{assumption}[Loss conditions]
\label{assump:loss}
Consider a loss function $\ell:\Rbb \to \Rbb_+$.
\begin{assumpenum}[leftmargin=*]
\item \label{assump:loss:convex} \textbf{Regularity.}~ 
Assume $\ell(\cdot)$ is continuously differentiable, convex, non-increasing, and $\ell(+\infty) = 0$.
Then the following function $\rho:[1,\infty)\to\Rbb_+$ is well-defined:
\begin{equation*}
    \rho(\lambda) := \min_{z\in \Rbb} \lambda \ell(z) +  z^2,\quad \lambda \ge 1.
\end{equation*}
\item \label{assump:loss:gradient-bound} \textbf{Lipschitzness.}~ 
Denote $g(\cdot) := |\ell'(\cdot)|$.
Assume there is a constant $C_g>0$ such that $g(\cdot) \le C_g $.
\item \label{assump:loss:local-smoothness} \textbf{Self-boundedness.}~
Assume there is a constant $C_{\beta}> 0$ such that $g(\cdot) \le C_{\beta} \ell(\cdot)$ and
\begin{align*}
 \ell(z) \le \ell(x) + \ell'(x) (z-x)  +C_\beta  g(x) (z-x)^2\ \ \text{for $z$ and $x$ such that} \ \ |z-x|< 1.
\end{align*}
\item \label{assump:loss:exp-tail} \textbf{An exponential tail.}~
Assume there is a constant $C_{e}>0$ such that
\(\ell(z) \le C_e  g(z)\) for $z\ge 0$.
\end{assumpenum}
\end{assumption}

The function $\rho$ in \Cref{assump:loss:convex} measures the squared length of the regularization path, which plays a central role in our bounds.
The second condition in \Cref{assump:loss:local-smoothness} is satisfied (with a possibly different constant) if $\ell$ is sufficiently differentiable and $\ell''(\cdot) \le C g(\cdot)$ for a constant $C>0$, hence it reflects the self-boundedness of $\ell$. 
Finally, one can verify that \Cref{assump:loss:convex,assump:loss:exp-tail} imply that $\ell(z) \le \ell(0) \exp(-C_e^{-1} z\big)$ for $z \ge 0$, that is, $\ell(\cdot)$ is exponentially tailed.

\Cref{prop:loss-examples} provides three examples of loss functions. The proof is included in \Cref{append:sec:loss-proof}.

\begin{proposition}[Examples]\label[proposition]{prop:loss-examples}
The following loss functions satisfy (parts of) \Cref{assump:loss}.
\begin{enumerate}[leftmargin=*]
\item
The logistic loss,
$\ell_{\log}(z) := \ln(1+\exp(-z))$, satisfies  \Cref{assump:loss} with $C_g=1$, $C_\beta=e/2$, $C_e=2$, and
\[ \rho(\lambda)\le  \rho_{\log}(\lambda) := 1+\ln^2(\lambda).\]

\item The ``flattened'' exponential loss with temperature $a>0$,
\begin{align*}
    \ell_{\exp}(z) := \begin{dcases}
        e^{-az} & z>0, \\
        1-az & z \le 0,
    \end{dcases}
\end{align*}
satisfies \Cref{assump:loss} with $C_g=a$, $C_\beta=\max\{a,\, a e^a/2,\, 1\}$, $C_e=1/a$, and 
\[ \rho(\lambda)\le \rho_{\exp}(\lambda):= 1+\ln^2(\lambda) / a^2.\]


\item The ``flattened'' polynomial loss of degree $a>0$,
\begin{align*}
    \ell_{\poly}(z) := \begin{dcases}
        (1+z)^{-a} & z> 0, \\
        1-a z & z\le 0,
    \end{dcases}
\end{align*}
satisfies \Cref{assump:loss:convex,assump:loss:gradient-bound,assump:loss:local-smoothness} with $C_g=a$, $C_\beta=\max\{a,\, (a+1)2^{a}\}$, and 
\[ \rho(\lambda)\le \rho_\poly(\lambda) := 2\lambda^{2/(a+2)}.\]
\end{enumerate}
\end{proposition}

The (flattened) polynomial loss is introduced in \citep{ji2020gradient,ji2021characterizing} and is later used by \citet{wang2022is} to improve importance weighting in distribution shift problems.
In the second and third examples, we flatten the negative parts of the exponential and polynomial losses to satisfy \Cref{assump:loss:gradient-bound}. 
This is necessary in our setting because, under (unflattened) exponential loss, large stepsize GD may oscillate catastrophically as shown by \citet{wu2023implicit}.


\paragraph{A two-layer network.}
We consider a two-layer network\footnote{We present our NTK results in a two-layer network for conciseness. Our results are ready to be extended to deep networks by combining our techniques with standard NTK arguments \citep{allen2019convergence,chen2020much}.} under a loss function $\ell$, defined as
\begin{equation}\label{eq:ntk:nn}
    L(\wB) := \frac{1}{n}\sum_{i=1}^n \ell \big(y_i f(\xB_i; \wB) \big),\quad 
    f(\xB; \wB) := \frac{1}{\sqrt{m}}\sum_{s=1}^m a_s \phi(\xB^\top \wB^{(s)}),\quad \phi(\cdot) := \max\{\cdot,0\},
\end{equation}
where the trainable parameters are denoted by
\(\wB \in \Rbb^{md}\), a stack of $\wB^{(1)},\dots, \wB^{(m)} \in \Rbb^d$.
We define $\phi'(0):=0$ for concreteness but our analysis can be easily generalized to other subderivatives, that is, $\phi'(0)=c$ for any $c\in [0,1]$.
Here, we follow the standard NTK setup   \citep{du2018gradient,ji2019polylogarithmic} and assume that $(a_s)_{s=1}^m$ are fixed parameters satisfying\footnote{Some papers assume that $(a_s)_{s=1}^m$ are uniformly sampled from $\{\pm 1\}$. In this case, $\big| \sum_{s=1}^m a_s \big| \le \sqrt{2m\ln (2/\delta)}$ with probability at least $1-\delta$. So we recover theirs by setting $C_a:=\sqrt{2\ln(2/\delta)}$ in \eqref{eq:ntk:a-condition} and applying a union bound.%
} 
\begin{equation}\label{eq:ntk:a-condition}
a_1,\dots, a_m \in \{\pm 1\},\quad 
    \bigg| \sum_{s=1}^m a_s \bigg| \le C_a \sqrt{m} \ \ \text{for some constant $C_a>0$}.
\end{equation}
It is clear that the optimization problem in \eqref{eq:ntk:nn} is non-convex.

We analyze the training of a nonlinear predictor (see $f$ in \eqref{eq:ntk:nn}) in the NTK regime \citep{jacot2018neural}.
The key idea is that, when the network width $m$ is large, the predictors induced by \eqref{eq:gd} iterates (in finite steps) are close to the gradient dynamics in a \emph{reproducing kernel Hilbert space} (RKHS), under the so-called \emph{neural tangent kernel} (NTK).
Therefore, the nonlinear predictors induced by the GD iterates are approximately linear in the NTK RKHS, allowing us to recycle our techniques for analyzing linear predictors in  \Cref{sec:logistic-reg}.

We reformulate our separability assumption within the NTK RKHS \citep{ji2019polylogarithmic}. 


\begin{assumption}[NTK separability]\label{assump:ntk:separable}
Assume the training data $(\xB_i, y_i)_{i=1}^n$ satisfies
\begin{assumpenum}
\item for every $i=1,\dots,n$, $\|\xB_i\|\le 1$ and $y_i \in \{\pm1\}$;
\item  there is a margin $\gamma>0$ and a map $\chi:\Rbb^d\to \Rbb^{d}$ such that $\|\chi(\cdot)\| \le 1$ and
\[
\min_{i=1,\dots,n} \Ebb_{\uB \sim \Ncal(0,\, \IB_d)}\big\la  y_i \phi'(\xB_i^\top \uB) \xB_i,\, \chi(\uB) \big\ra \ge \gamma ,
\]
that is, the dataset is separable in the NTK RKHS space.
\end{assumpenum}
\end{assumption}

\Cref{assump:ntk:separable} corresponds to Assumption 2.1 in \citep{ji2019polylogarithmic}.
\Cref{assump:ntk:separable} requires the dataset being linearly separable under the infinite-dimensional NTK feature, that is, $\phi'(\xB^\top \uB)\xB$ indexed by $\uB$, which corresponds to the gradient of $f(\xB; \cdot)$ at the Gaussian random initialization as the network width $m\to\infty$ (ignoring $\frac{1}{\sqrt{m}}a_s$'s for simplicity).
\Cref{assump:ntk:separable} is satisfied when the dataset is linearly separable (as in \Cref{assump:sgd:bounded-separable}, with a possibly different margin parameter).
In addition, \Cref{assump:ntk:separable} can capture non-linearly separable datasets.
We refer the reader to Section 5 in \citep{ji2019polylogarithmic} for more discussion.

We are ready to state our theorem for general loss functions and two-layer networks.

\begin{theorem}[General losses and NTK]\label{thm:ntk}
Consider \cref{eq:gd} with stepsize $\eta>0$ for learning a two-layer network \cref{eq:ntk:nn} under a loss function $\ell$ that satisfies \Cref{assump:loss:convex,assump:loss:gradient-bound}.
Suppose \eqref{eq:ntk:a-condition} and \Cref{assump:ntk:separable} hold, and the network is initialized by 
\[
\wB_{0} \sim \Ncal(0,\, \IB_{md}).
\]
Let $T$ be the maximum number of steps.
Suppose the network width $m$ is at least
\begin{align*}
    m\ge  \bigg(\frac{30R^{1/3} + 10\ln^{1/4}(n/\delta)}{\gamma}\bigg)^6,\quad 
    \text{where}\ \ 
    R := 6\frac{ \sqrt{\rho(\gamma^2 \eta T)}  + C_a+\sqrt{2\ln(2n/\delta)}  +  \eta  C_g}{\gamma}.
\end{align*}
Then with probability at least $1-3\delta$ over the randomness of initialization, the following holds:
\begin{itemize}[leftmargin=*]
\item \textbf{Lazy training.}~
For every $t \le T$, we have 
\[
\|\wB_t - \wB_0\|\le R.
\]

\item \textbf{The EoS phase.}~
For every $t \le T$ (and in particular in the EoS phase), we have 
\begin{align*}
\frac{1}{t}\sum_{k=0}^{t-1} L(\wB_k) \le 9\frac{\rho(\gamma^2 \eta t) + \big(C_a+\sqrt{2\ln(2n/\delta)}  + \eta C_g \big)^2 }{\gamma^2 \eta t}.
\end{align*}

\item \textbf{The stable phase.}~ 
Assume the loss $\ell$ also satisfies \Cref{assump:loss:local-smoothness}.
If $s<T$ is such that 
\begin{equation}\label{eq:ntk:stable-phase-criteria}
    L(\wB_s) \le \min\bigg\{ \frac{1}{ 12 C_\beta^2 \eta} ,\; \frac{\ell(0)}{n}\bigg\},
\end{equation}
then \eqref{eq:gd} is in the stable phase, that is, 
$\big(L(\wB_t)\big)_{t=s}^T$ decreases monotonically, and moreover,
\begin{align*}
    L(\wB_t) \le 15 \frac{\rho (\gamma^2 \eta (t-s)) }{\gamma^2 \eta (t-s)},\quad t\in (s,\, T].
\end{align*}
\item \textbf{Phase transition time.}~
There exists a constant $C_1>0$ that only depends on $C_g$, $C_\beta$, $C_a$, $\ell(0)$, and $\ln(1/\delta)$ such that the following holds. 
Let 
\begin{equation*}
    \tau := \frac{1}{\gamma^2 } \max\bigg\{ \frac{\psi^{-1}\big(C_1(\eta + n) \big)}{\eta},\ C_1 (\eta + n)\eta  \bigg\},\ \ \text{where}\ \  \psi(\lambda) := \frac{\lambda}{\rho(\lambda)}.
\end{equation*}
If $\tau \le T$, then \cref{eq:ntk:stable-phase-criteria} holds for some $s\le \tau$.

\item \textbf{Phase transition time under an exponential tail.}~
Assume the loss $\ell$ further satisfies  \Cref{assump:loss:exp-tail}, then there exists a constant $C_2>0$ that only depends on $C_e$, $C_g$, $C_\beta$, $C_a$, $\ell(0)$, and $\ln(1/\delta)$ such that the following holds. 
Let
\begin{equation*}
    \tau := \frac{C_2}{\gamma^2} \max\big\{ \eta, \ n\ln(n) \big\}.
\end{equation*}
If $\tau \le T$, then \cref{eq:ntk:stable-phase-criteria} holds for some $s\le \tau$. 
\end{itemize}
\end{theorem}

The proof of \Cref{thm:ntk} is deferred to \Cref{append:sec:ntk-proof}.
\Cref{thm:ntk} characterizes the behaviors of large stepsize GD for a wide two-layer network under general classification losses. 
We remark that the proof of \Cref{thm:ntk} directly adapts when replacing the two-layer network with a linear predictor. 
For completeness, we state a variant of \Cref{thm:ntk}
for GD in linear classification under general losses as \Cref{thm:general-loss} in \Cref{append:sec:general-loss}.

\paragraph{Convergence rate.}
The final convergence rate of GD in \Cref{thm:ntk} depends on the loss functions and the stepsizes. 
We summarize and compare several key examples in \Cref{tab:ntk},
where we assume the total number of steps $T$ is large and treat all other 
instance specific
quantities (such as $n$ and $\gamma$) as constants. 
Results in \Cref{tab:ntk} are direct consequences of  \Cref{thm:ntk} and \Cref{prop:loss-examples}.
In particular, we see that GD attains an improved loss by using a large stepsize that balances the length of the EoS and the stable phases. 

\setcellgapes{2pt}
\begin{table}[t]
\centering
\makegapedcells
\caption{Effects of the loss functions and stepsizes in \Cref{thm:ntk}.}
\label{tab:ntk}
\begin{tabular}{c|cc|ccc}
\toprule
loss function & \multicolumn{2}{c|}{logistic / flattened exponential} & \multicolumn{3}{c}{flattened polynomial of degree $a$}  \\
\midrule
degree condition  & \multicolumn{2}{c|}{N/A} &  $a>0$ & $0<a\le 1$ & $a>1$ \\
\midrule
$\rho(\lambda)$ & \multicolumn{2}{c|}{$ \Theta\big( \ln^2(\lambda) \big)$} & \multicolumn{3}{c}{$\Theta\big( \lambda^{\frac{2}{a+2}} \big)$}  \\
\midrule
stepsize $\eta$   & $1$ & $\Theta(T)$ & $1$ & $\Theta(T^{\frac{a}{2}})$ & $\Theta(T^{\frac{1}{2}})$ \\
width $m$     &  $\Omega\big(\ln^2(T)\big) $ & $\Omega(T^2)$  & $\Omega\big(T^{\frac{2}{a+2}}\big)$ & $\Omega(T)$ & $\Omega(T)$  \\
\makecell{phase transition\\ time $s$}  &  N/A & $\le T/2$ & N/A & $\le T/2$ & $\le T/2$ \\
loss $L(\wB_T)$    & $\Ocal\big({\ln^2(T)}/{T}\big)$  & $\Ocal\big({\ln^2(T)}/{T^2}\big)$ & $\Ocal\big(T^{\frac{-a}{a+2}}\big)$ & $\Ocal(T^{-\frac{a}{2}})$ & $\Ocal\big(T^{\frac{-3a}{2a+4}}\big)$ \\
\bottomrule
\end{tabular}
\end{table}

\paragraph{Loss conditions.}
Notably, \Cref{thm:ntk} (also \Cref{thm:general-loss} in \Cref{append:sec:general-loss}) reflects the role of each loss function condition in \Cref{assump:loss}.
Specifically, a general classification loss is specified by \Cref{assump:loss:convex}. 
Then the EoS phase bound in \Cref{thm:ntk} holds when the loss satisfies an additional Lipschitz condition (\Cref{assump:loss:gradient-bound}), and the stable phase bound holds if the loss further satisfies the self-boundedness condition (\Cref{assump:loss:local-smoothness}).
Finally, an exponential tail condition (\Cref{assump:loss:exp-tail}) implies a better phase transition bound.

\paragraph{The width condition.}
The width condition in \Cref{thm:ntk} depends on the loss function $\ell$ and the stepsize $\eta$.
When we specialize $\ell$ to the logistic loss (so $\rho(\lambda) = \Ocal(\ln^2(\lambda))$ by \Cref{prop:loss-examples})
and the stepsize to a constant
($\eta = \Theta(1)$), 
the width condition is \emph{polylogarithmic} in the number of samples and the number of iterates, and the achieved convergence rate is $\tilde\Ocal(1/t)$. 
This recovers the main result of \citet{ji2019polylogarithmic}.

On the other hand, the width condition in \Cref{thm:ntk} is at least $\Omega(\eta^2)$, which will become \emph{polynomial} in the number of steps when GD is equipped with a polynomially large stepsize $\eta = \poly(T)$.
This is in stark contrast to the previous example where $\eta = \Theta(1)$ and the width only needs to be polylogarithmic in $T$,  suggesting that a larger stepsize helps GD escape the NTK regime in the early optimization phase. 
We leave it as a future direction to study the early-phase feature learning of GD caused by a large stepsize.

The loss function affects the width condition in \Cref{thm:ntk} through $\rho(\gamma^2 \eta T)$. 
For instance, the width condition becomes $m\ge \Omega(T^{2/(a+2)})$ for the constant stepsize and the flattened polynomial loss of degree $a>0$ (see \Cref{tab:ntk}). Such results are new in the NTK literature to our knowledge.

\section{Conclusion}\label{sec:conclusion}
We study constant-stepsize GD for training linear and nonlinear predictors (in the NTK regime) under general classification loss functions and constant-stepsize SGD for logistic regression,
assuming suitable separability conditions on the dataset. 
We show GD and SGD converge even with a large stepsize that leads to a locally oscillatory loss.
Moreover, we show that a large stepsize allows GD to attain an accelerated loss by undergoing an unstable initial phase, which cannot be attained if the stepsize is small such that GD converges monotonically. 
We leave for future work to relax the separability conditions. 







\acks{
We thank the anonymous reviewers and area chairs for their helpful comments.
We thank Fabian Pedregosa for his suggestions on an early draft.
We gratefully acknowledge the support of the NSF and the Simons Foundation for the Collaboration on the Theoretical Foundations of Deep Learning through awards DMS-2031883 and \#814639 respectively, NSF Grant DMS 2015341 and NSF grant 2023505 on Collaborative Research: Foundations of Data Science Institute (FODSI).}

\bibliography{ref}
\newpage
\appendix 

\section{Additional Simulation}\label{append:sec:additional-exp}
\begin{figure}
    \centering
    \includegraphics[width=0.5\linewidth]{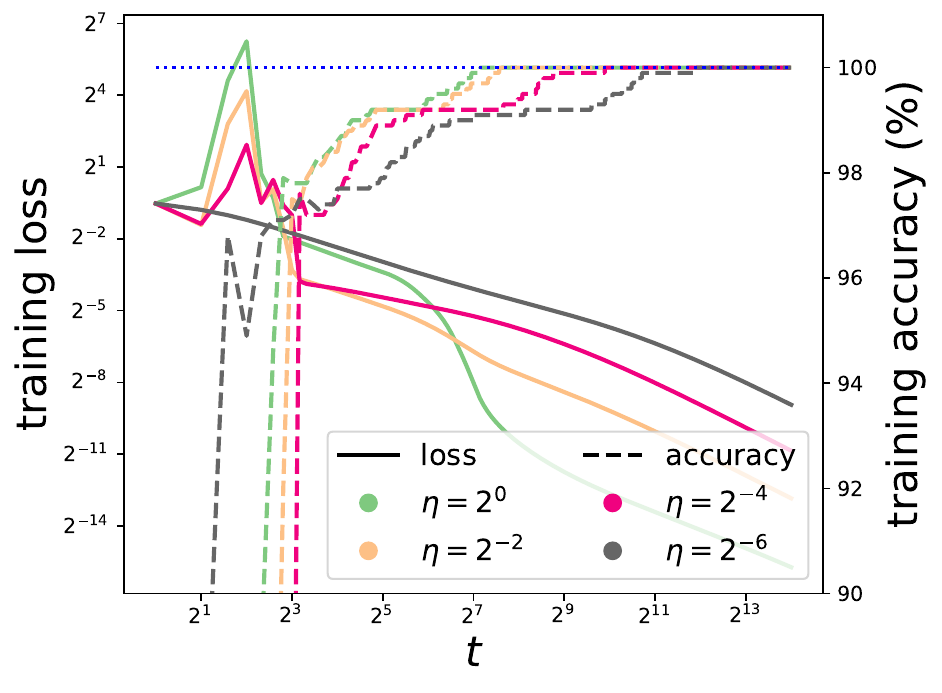}
    \caption{Training loss and training accuracy of \Cref{eq:gd} for logistic regression on a subset of MNIST. This is a replication of \Cref{fig:mnist} with training accuracy curves. The plots show that the stable phase is entered before reaching $100\%$ training accuracy. Moreover, GD with larger stepsizes reaches perfect training accuracy faster.}
    \label{fig:mnist-acc}
\end{figure}

In \Cref{fig:mnist-acc}, we recreate \Cref{fig:mnist} with training accuracy curves. The plots show that the stable phase is entered before reaching $100\%$ training accuracy. Moreover, GD with larger stepsizes reaches perfect training accuracy faster.
The simulations suggest that the phase transition time bound given in \Cref{thm:logistic-reg}, especially the dependence on the sample size $n$, might be improvable. 

\section{Proof of Theorem \ref{thm:logistic-reg}}\label{append:sec:logistic-proof}
Throughout this part, we assume \Cref{assump:bounded-separable} holds. In addition, without loss of generality, we assume $y=1$.
We define
\[
L(\wB) = \frac{1}{n}\sum_{i=1}^n \ell (\xB^\top_i\wB), \quad 
\ell(t) = \ln(1+e^{-t}),\quad 
G(\wB) =\frac{1}{n}\sum_{i=1}^n |\ell'(\xB^\top_i\wB)|.
\]

The following lemma presents a new split optimization method for handling large stepsizes. 
\begin{lemma}[A split optimization bound]\label[lemma]{lemma:implicit-bias}
For every $\eta>0$, $\wB_0$, and $\uB = \uB_1 + \uB_2$ such that 
\[
\uB_2 = \frac{\eta }{2\gamma} \wB_*,
\]
we have 
\begin{align*}
    \frac{\|\wB_t - \uB\|^2}{2\eta t} + \frac{1}{t} \sum_{k=0}^{t-1} L(\wB_k) \le L(\uB_1) + \frac{\|\wB_0 - \uB\|^2}{2\eta t},\quad t\ge 1.
\end{align*}
\end{lemma}
\begin{proofof}[\Cref{lemma:implicit-bias}]
Using the definition of GD, we have
\begin{align*}
    \|\wB_{t+1} - \uB \|^2 
    &= \| \wB_t-\uB\|^2 + 2\eta \la \grad L(\wB_t) , \uB -\wB_t \ra + \eta^2\big\|\grad L(\wB_t) \big\|^2_2 \\
     &= \| \wB_t-\uB\|^2 + 2\eta \la \grad L(\wB_t) , \uB_1-\wB_t \ra  + \eta \Big( 2\la \grad L(\wB_t) ,  \uB_2 \ra + \eta\big\|\grad L(\wB_t) \big\|^2_2 \Big).
 \end{align*} 
Plugging $\uB_2$ and $\grad L$, we can show the second term is always non-positive:
\begin{align*}
   &\quad \ 2 \la \grad L(\wB_t) ,  \uB_2 \ra + \eta \big\|\grad L(\wB_t) \big\|^2_2 \\
   &= \frac{2}{n }\sum_{i=1}^n \ell'(\wB^\top_t \xB_i) \la \xB_i, \uB_2 \ra + \eta\bigg\|\frac{1}{n} \sum_{i=1}^n \ell'(\wB^\top_t \xB_i) \xB_i \bigg\|^2_2 \\
   &\le \frac{2\gamma \|\uB_2\|}{ n }\sum_{i=1}^n \ell'(\wB^\top_t \xB_i)  + \eta \bigg(\frac{1}{n} \sum_{i=1}^n \ell'(\wB^\top_t \xB_i) \bigg)^2  && \explain{by \(\xB_i^\top\wB_* \ge \gamma\) and \(\|\xB_i\|\le1\)}\\
   &\le (-{2\gamma\|\uB_2\|}+ \eta ) \cdot \frac{1}{ n }\sum_{i=1}^n |\ell'(\wB_t^\top \xB_i)| && \explain{since $|\ell'|\le 1$} \\
   &\le 0. && \explain{by the choice of \(\uB_2\)}
\end{align*}
So we have 
\begin{align*}
    \|\wB_{t+1} - \uB \|^2 
    &\le \| \wB_t-\uB\|^2 + 2\eta \la \grad L(\wB_t),  \uB_1- \wB_t \ra  \\
    &\le  \| \wB_t-\uB\|^2 + 2\eta \big( L(\uB_1) - L( \wB_t) \big). && \explain{by convexity}
\end{align*} 
Telescoping the sum from $0$ to $t$, we obtain
\begin{align*}
   \frac{  \|\wB_t - \uB \|^2}{2\eta} + \sum_{k=0}^{t-1} L(\wB_k) 
   &\le t L(\uB_1) + \frac{ \| \wB_0 - \uB\|^2 }{2\eta},
\end{align*}
which completes the proof.
\end{proofof}

Applying the above lemma with an appropriate comparator, we can get the following bounds on the risk and parameter norm.
\begin{lemma}[Parameter and risk bounds in the EoS phase]\label[lemma]{lemma:parameter-risk-bound}
Let $\wB_0=0$. For every $\eta>0$, we have
\begin{align*}
    \frac{1}{t}\sum_{k=0}^{t-1} L(\wB_k) \le \frac{1 + \ln^2(\gamma^2 \eta t) + \eta^2/4}{ \gamma^2 \eta t},\quad
    \|\wB_t\| \le \frac{\sqrt{2} +2\ln(\gamma^2 \eta t) + \eta}{\gamma},\quad t\ge 1.
\end{align*}
\end{lemma}
\begin{proofof}[\Cref{lemma:parameter-risk-bound}]
In \Cref{lemma:implicit-bias}, choose
\[
\wB_0 = 0,\quad \uB_1 = \frac{\ln(\gamma^2 \eta t)}{\gamma} \wB_*,
\]
and check that 
\begin{align*}
    L(\uB_1) &\le \frac{1}{n} \sum_{i=1}^n \exp(- \xB_i^\top \uB_1) \\
    &\le \exp( -\gamma\|\uB_1\| ) \\
    &\le \frac{1}{\gamma^2 \eta t},
\end{align*}
and that 
\begin{align*}
    \|\uB\| = \|\uB_1 +\uB_2\|
    = \frac{\ln(\gamma^2 \eta t) + \eta/2}{\gamma}.
\end{align*}
We then apply \Cref{lemma:implicit-bias} to get
\begin{align*}
    \frac{1}{t}\sum_{k=0}^{t-1} L(\wB_k) \le L(\uB_1) + \frac{\|\uB\|^2}{2\eta t} \le \frac{1 + \ln^2(\gamma^2 \eta t) + \eta^2/4}{ \gamma^2 \eta t},
\end{align*}
and 
\begin{align*}
    \|\wB_t\| \le \|\wB_t -\uB\| + \|\uB\| \le \sqrt{2\eta t L(\uB_1) } +  2\|\uB\|\le \frac{\sqrt{2} +2\ln(\gamma^2 \eta t) + \eta}{\gamma}.
\end{align*}
These complete the proof.
\end{proofof}

The following lemma controls the gradient potential. 
We remark that the bound on gradient potential in \Cref{lemma:gradient-bound} does not scale with $\eta$, while the bound on risk in \Cref{lemma:parameter-risk-bound} linearly scales with $\eta$.
This difference will be crucial for getting a sharp bound on the phase transition time. 
\begin{lemma}[Gradient potential bound in the EoS phase]\label[lemma]{lemma:gradient-bound}
Let $\wB_0 = 0$.
For every $\eta>0$, we have
\begin{align*}
    \frac{1}{t} \sum_{k=0}^{t-1} G(\wB_k)  \le \frac{\sqrt{2} + 2\ln(\gamma^2 \eta t) + \eta}{\gamma^2 \eta t},\quad t\ge 1.
\end{align*}
\end{lemma}
\begin{proofof}[\Cref{lemma:gradient-bound}]
This is from the perceptron argument \citep{novikoff1962convergence}. Specifically, 
\begin{align*}
    \la \wB_{t+1}, \wB_*\ra &= \la \wB_t, \wB_*\ra  - \eta \la \grad L(\wB_t), \wB_*\ra  \\
    &= \la \wB_t, \wB_*\ra  - \frac{\eta}{n} \sum_{i=1}^n \ell'(\xB_i^\top \wB_t) \la \xB_i, \wB_*\ra \\ 
    &\ge  \la \wB_t, \wB_*\ra - \frac{\gamma \eta}{n} \sum_{i=1}^n \ell'(\xB_i^\top \wB_t) \\
    &=  \la \wB_t, \wB_*\ra + \gamma \eta G(\wB_t).
\end{align*}
So we have 
\begin{align*}
      \frac{1}{t} \sum_{k=0}^{t-1} G(\wB_k) \le \frac{\la \wB_t, \wB_*\ra - \la \wB_0, \wB_*\ra }{\gamma \eta t}\le \frac{\|\wB_t - \wB_0\|}{\gamma \eta t}.
\end{align*}
We complete the proof by applying the parameter bound in \Cref{lemma:parameter-risk-bound}.
\end{proofof}

For logistic regression, the local sharpness is related to the loss value. So the landscape will become sufficiently flat when the loss is sufficiently small, allowing GD to fall into the stable convergence phase.
Our next lemma formally justifies this discussion.
\begin{lemma}[Stable phase]\label[lemma]{lemma:stable-phase}
If there exists $s \ge 0$ such that
\[
L(\wB_s) \le \frac{2}{ \eta},
\]
then for every $t\ge s$,  $L(\wB_t)$ is non-increasing.
\end{lemma}
\begin{proofof}[\Cref{lemma:stable-phase}]
We verify that $\ln L(\wB)$ is $1$-smooth (this is used by \citet{ji2018risk}) by direct computation:
\begin{align*}
    \frac{\dif^2}{(\dif \wB)^2} \ln L(\wB) &= \frac{1}{L(\wB)^2} \Big( \grad^2 L(\wB) \cdot L(\wB) - \big(\grad L(\wB)\big)^{\otimes 2} \Big) \\
    &\preceq \frac{1}{L(\wB)} \grad^2 L(\wB)  \\
    &\preceq \frac{1}{L(\wB)} L(\wB)\cdot  \IB = \IB. && \explain{since \(|\ell''| \le \ell\) and \(\|\xB_i\|\le 1\)}
\end{align*}
Using the $1$-smoothness of $\ln L(\cdot)$, we get
\begin{align*}
    \ln L(\wB_{t+1}) &\le \ln L(\wB_t) + \bigg\la \frac{\grad L(\wB_t)}{L(\wB_t)}, \wB_{t+1}- \wB_t \bigg\ra + \frac{1}{2} \|\wB_{t+1} - \wB_t\|^2 \\
    &= \ln L(\wB_t) - \frac{\eta }{L(\wB_t)} \|\grad L(\wB_t)\|^2 + \frac{\eta^2}{2} \|\grad L(\wB_t)\|^2 \\
    &= \ln L(\wB_t) - \eta  \bigg( \frac{1}{L(\wB_t)} - \frac{\eta}{2}\bigg) \|\grad L(\wB_t)\|^2 .
\end{align*}
The above implies 
\[
L(\wB_{t+1}) \le L(\wB_t) \  \ \text{if} \ \ L(\wB_t) \le \frac{2}{\eta}.
\]
We complete the proof by induction and the condition that $L(\wB_s)\le 2/\eta$.
\end{proofof}

The next lemma provides a risk bound for GD in the stable phase. 
Notably, the bound depends on the initial condition ($\wB_s$) through an exponential potential (see the definition of $F(\cdot)$) instead of a quadratic potential. We will see later that $F(\wB_s)$ can be made a constant while $\|\wB_s\|^2$ might be as large as $\Theta(\eta^2)$.
\begin{lemma}[A risk bound in the stable phase]\label[lemma]{lemma:stable-phase:risk-bound}
Suppose there exists a time $s$ such that 
\begin{align*}
    L(\wB_s) \le \frac{1}{ \eta}.
\end{align*}
Then for every $t \ge s$, we have 
\begin{align*}
    L(\wB_t) \le \frac{2F(\wB_s) + \ln^2 (\gamma^2 \eta (t-s)) }{\gamma^2 \eta (t-s)},\ \  \text{where}\ \ 
F(\wB) := \frac{1}{n} \sum_{i=1}^n \exp(-\xB_i^\top \wB).
\end{align*}
\end{lemma}
\begin{proofof}[\Cref{lemma:stable-phase:risk-bound}]
The assumption enables \Cref{lemma:stable-phase} for all $t\ge s$.
Therefore we have 
\begin{align*}
\text{for every $t\ge s$},\quad 
    L(\wB_{t}) 
    \le L(\wB_{t-1})
    \le \cdots 
    \le L(\wB_s)\le \frac{1}{ \eta}.
\end{align*}
Because the logistic loss with $\|\xB\|\le 1$ satisfies $\|\grad L(\cdot)\| \le L(\cdot)$,
for a comparator $\uB$ and $t\ge s$, we have 
 \begin{align*}
     \|\wB_{t+1} - \uB\|^2 &= \|\wB_t - \uB\|^2 + 2\eta \la \grad L(\wB_t), \uB - \wB_t \ra + \eta^2 \|\grad L(\wB_t)\|^2 \\
     &\le \|\wB_t - \uB\|^2 + 2\eta \big( L(\uB) - L(\wB_t) \big) + \eta^2 L(\wB_t)^2 && \explain{since \(|\ell'| \le \ell\)} \\
     &\le \|\wB_t - \uB\|^2 + 2\eta \big( L(\uB) - L(\wB_t) \big) + \eta L(\wB_t) && \explain{since \(L(\wB_t) \le 1/\eta\)} \\
     &=  \|\wB_t - \uB\|^2 + 2\eta  L(\uB) - \eta L(\wB_{t}).
 \end{align*}  
Telescoping the sum from $s$ to $t$, we get 
\begin{align*}
    \|\wB_{t} - \uB \|^2 + \eta \sum_{k=s}^{t-1} L(\wB_k) \le \|\wB_{s} - \uB\|^2 + 2\eta (t-s) L(\uB),
\end{align*}
that is,
\begin{align*}
    \frac{1}{t-s}\sum_{k=s}^{t-1} L(\wB_{k}) &\le 2L(\uB) + \frac{\|\wB_s - \uB\|^2  - \|\wB_t -\uB\|^2}{\eta (t-s)} \\
    &\le 2 L(\uB) + \frac{\|\wB_s - \uB\|^2}{\eta (t-s)} .
\end{align*}
Choose 
\[
\uB = \wB_{s} + \uB_1, \quad \uB_1 = \frac{\ln\big(\gamma^2 \eta (t-s) \big)}{\gamma} \wB_*,  
\]
and verify that 
\begin{align*}
    \|\wB_s - \uB\| = \frac{\ln\big(\gamma^2 \eta (t-s) \big)}{\gamma},
\end{align*}
and
\begin{align*}
    L(\uB) &\le F(\uB) = \frac{1}{n} \sum_{i=1}^n \exp(- \la \uB, \xB\ra ) \\
    &= \frac{1}{n} \sum_{i=1}^n \exp(- \la \wB_{s}, \xB\ra - \la \uB_1,\xB\ra) \\
    &\le \frac{1}{n} \sum_{i=1}^n \exp(- \la \wB_{s}, \xB\ra ) \cdot \frac{1}{\gamma^2 \eta (t-s)} \\
    &= \frac{ F(\wB_{s})}{\gamma^2 \eta (t-s)}.
\end{align*}
Bringing these back, we obtain
\begin{align*}
    \frac{1}{t-s}\sum_{k=1}^{t-s} L(\wB_{s+k}) 
    \le 2L(\uB) + \frac{\|\wB_s - \uB\|^2}{\eta (t-s)} 
    \le   \frac{2F(\wB_s)+\ln^2\big(\gamma^2 \eta (t-s) \big)}{\gamma^2 \eta (t-s)} .
\end{align*}
We complete the proof using the monotonicity of $L(\wB_t)$ for $t\ge s$ by \Cref{lemma:stable-phase}.
\end{proofof}

The following lemma provides a bound on the phase transition time. 
The bound is linear in $\eta$ because we use the gradient potential bound in \Cref{lemma:gradient-bound} instead of the risk bound in \Cref{lemma:parameter-risk-bound}.
\begin{lemma}[Phase transition time]\label[lemma]{lemma:phase-transition}
Let $\wB_0 = 0$.
For every $\eta > 0$, 
let 
\begin{equation*}
    \tau := \frac{60}{\gamma^2}\max\bigg\{\eta,\ n,\ e,\ \frac{\eta + n}{\eta}\ln\frac{\eta + n}{\eta} \bigg\}.
\end{equation*}
Then there exists $0\le s \le \tau$ such that 
\begin{align*}
    L(\wB_s) \le \frac{1}{ \eta },\quad F(\wB_s) \le 1.
\end{align*}
\end{lemma}
\begin{proofof}[\Cref{lemma:phase-transition}]
Applying \Cref{lemma:gradient-bound} with $t=\tau$, we have
\begin{align*}
    \frac{1}{\tau} \sum_{k=0}^{\tau-1} G(\wB_k)  &\le \frac{\sqrt{2} + 2\ln(\gamma^2 \eta \tau ) + \eta}{\gamma^2 \eta \tau } \\ 
    &\le \frac{\sqrt{2} + 2\ln( \gamma^2\tau ) + 3\eta}{\eta \gamma^2  \tau } && \explain{since \(\ln(\eta)\le \eta\)} \\ 
    &\le \frac{1}{2(\eta + n)} \le  \min\bigg\{\frac{1}{2 \eta},\;  \frac{1}{2n} \bigg\},
\end{align*}
where the third inequality is by verifying that 
\begin{align*}
    \frac{\sqrt{2} + 2\ln(\gamma^2\tau)}{\gamma^2 \tau} \le \frac{\eta}{4(\eta + n)} \quad &\text{if} \quad \gamma^2 \tau \ge 60 \frac{\eta + n}{\eta} \max\bigg\{\ln \frac{\eta + n}{\eta},\ 1\bigg\}, \\ 
    \frac{3}{\gamma^2 \tau} \le \frac{1}{4(\eta + n)} \quad &\text{if} \quad \gamma^2 \tau \ge 12(\eta + n),
\end{align*}
and that the two conditions are satisfied because 
\begin{align*}
    \gamma^2 \tau := 60\max\bigg\{\eta,\ n,\ e, \frac{\eta + n}{\eta} \ln \frac{\eta + n}{\eta}\bigg\}
    \ge \max\bigg\{ 12(\eta + n), \  60 \frac{\eta + n}{\eta} \max\bigg\{\ln \frac{\eta + n}{\eta}, 1\bigg\} \bigg\}.
\end{align*}
So there exists $s \le \tau$ such that 
\[
G(\wB_s) \le  \min\bigg\{\frac{1}{2 \eta},\;  \frac{1}{2n} \bigg\}.
\]
The second bound ensures that 
\begin{align*}
\text{for every $i$},\quad 
   \frac{1}{n} \cdot \frac{1}{1+\exp(\xB_i^\top \wB_s)} \le G(\wB_s) \le \frac{1}{2n},
\end{align*}
that is,
\begin{align*}
\text{for every $i$},\quad 
    \xB_i^\top \wB_s \ge 0.
\end{align*}
So we have
\begin{align*}
F(\wB_s) = \frac{1}{n}\sum_{i=1}^n \frac{2}{2\exp(\xB^\top_i\wB_s)}
\le \frac{1}{n}\sum_{i=1}^n \frac{2}{1+\exp(\xB^\top_i\wB_s)} = 2 G(\wB_s) \le \min\bigg\{\frac{1}{  \eta},\;  \frac{1}{n} \bigg\}.
\end{align*}
Furthermore, by definitions of $L$ and $F$ we have
\begin{align*}
    L(\wB_s) \le F(\wB_s) \le \min\bigg\{ \frac{1}{   \eta},\, \frac{1}{n} \bigg\} \le 1,
\end{align*}
which completes the proof.
\end{proofof}

The above lemmas imply \Cref{thm:logistic-reg}.
\begin{proofof}[\Cref{thm:logistic-reg}]
It follows from \Cref{lemma:parameter-risk-bound,lemma:stable-phase:risk-bound,lemma:phase-transition}.
\end{proofof}

\section{Proof of Corollary \ref{cor:acceleration}}\label{append:sec:acceleration}
\begin{proofof}[\Cref{cor:acceleration}]
We first verify $\tau \le T/2$ for 
\begin{align*}
\tau &:= \frac{60}{\gamma^2}\max\bigg\{\eta,\ n,\ e,\ \frac{\eta + n}{\eta}\ln\frac{\eta + n}{\eta} \bigg\}.
\end{align*}
The first term is 
\begin{align*}
    \frac{60}{\gamma^2} \eta 
    &= \frac{T}{2}, && \explain{since \(\eta := \frac{\gamma^2}{120}T\)}
\end{align*}
the second term is 
\begin{align*}
    \frac{60n}{\gamma^2} \le \frac{T}{2}, && \explain{since \(T \ge \frac{120 \max\{e,n\}}{\gamma^2}\)},
\end{align*}
the last term is 
\begin{align*}
     \frac{60e}{\gamma^2} \le \frac{T}{2}, && \explain{since \(T \ge \frac{120 \max\{e,n\}}{\gamma^2}\)},
\end{align*}
and the third term is less than $2\ln(2) \le T/2$ (note that $T\ge 120$ since $\gamma<1$ and $n\ge 1$) because 
\begin{align*}
\frac{\eta + n}{\eta} 
&= \frac{T\gamma^2 / 120 + n}{T\gamma^2/120} && \explain{since \(\eta := \frac{\gamma^2}{120}T\)} \\ 
&\le  \frac{T\gamma^2 / 120 + T\gamma^2 / 120}{T\gamma^2/120} && \explain{since \(T \ge \frac{120 n}{\gamma^2}\)} \\
&= 2.
\end{align*}
We have verified that $\tau \le T/2$.

Next, we apply \Cref{thm:logistic-reg} with $s = \tau \le T/2$ and get 
\begin{align*}
L(\wB_T) 
&\le \frac{2F(\wB_s) + \ln^2 (\gamma^2 \eta (T-s)) }{\gamma^2 \eta (T-s)} \\
&\le \frac{2 + \ln^2 (\gamma^2 \eta T/2) }{\gamma^2 \eta T/2} && \explain{since \(F(\wB_s) \le 1\) and \(s\le T/2\)} \\
&= \frac{2 + \ln^2 (\gamma^4 T^2/240) }{\gamma^4 T^2/240} && \explain{since \(\eta := \frac{\gamma^2}{120}T\)}\\
&\le \frac{480 \ln^2(\gamma^4 T^2)}{\gamma^4 T^2},
\end{align*}
which completes the proof.
\end{proofof}

\section{Proof of Theorem \ref{thm:gd:lb}}\label{append:sec:lower-bound}
The following lemma suggests an $\Omega(1/(\eta t))$ lower bound on the risk for GD with a fixed stepsize.
\begin{lemma}[A lower bound]\label[lemma]{lemma:lb:asymp-rate}
Suppose that $(\xB_i, y_i)_{i=1}^n$ are linearly separable with max-margin direction $\wB_*$ and margin $\gamma$.
Denote 
\[
\bar \xB_i = \xB_i - \la \xB_i, \wB_*\ra \wB_*, \quad i=1,\dots,n.
\]
Assume that $(\bar \xB_i, y_i)_{i=1}^n$ are non-separable, that is, there exists $b>0$ such that
\begin{align*}
    \sup_{\bar \vB\in \Rbb^{d-1}, \|\bar \vB\|=1}\min_{i}\la y_i \bar \xB_i, \bar \vB \ra \le -b.
\end{align*}
Then for every stepsize $\eta>0$, we have
\[L(\wB_t) \ge\frac{1}{c_0\eta \exp(c_0 \eta) t},\quad t\ge 1,\]
where $c_0>0$ is a function of $\gamma, b$ and $n$ but is independent of $\eta$ and $t$.
\end{lemma}
\begin{proofof}[\Cref{lemma:lb:asymp-rate}]
Assume $y_i=1$ without loss of generality.
Define 
\begin{align*}
    w_t = \la \wB_t, \wB_*\ra,\quad 
    \bar \wB_t = \wB_t - \la \wB_t, \wB_*\ra \wB_*.
\end{align*}
We call some results in  \citep{wu2023implicit}.
By Lemma B.3 in  \citep{wu2023implicit} (replacing their $\eta n$ by $\eta$ since they do not take average in their loss definition), we get 
\begin{align*}
        \|\bar\wB_t\| \le \max\{4n/b, \eta n/b\} + \eta \le c_1 \eta,
\end{align*}
where $c_1$ is a positive constant independent of $\eta$.
By Lemma B.7 in  \citep{wu2023implicit}, we get 
\begin{align*}
     w_t &\le \frac{1}{\gamma}\ln \bigg( \bigg( e \frac{\eta}{n} \gamma^2 G_{\max}+e \frac{\eta}{n} \gamma H_{\max}\bigg) (t+1) \bigg) ,\quad  t\ge 0,
\end{align*}
where $G_{\max}$ and $H_{\max}$ can be upper bounded by (see their Definition 3 in Appendix B.3)
\begin{align*}
    G_{\max} + H_{\max} &=\sup_{\|\bar \wB_t\|\le c_1 \eta}  \sum_{i=1}^n \exp(- \bar \xB_i^\top \bar \wB  ) \le n \exp( c_1 \eta). 
\end{align*}
So we have 
\begin{align*}
    w_t \le \frac{1}{\gamma} \ln \big( e \gamma \eta \exp(c_1 \eta) (t+1) \big),\quad t\ge 0.
\end{align*}
The above implies that
\begin{align*}
    L(\wB_t) &= \frac{1}{n}\sum_{i=1}^n \ln (1+\exp(-\xB_i^\top \wB_t)) \\
    &=\frac{1}{n}\sum_{i=1}^n  \ln \big(1+ \exp(-\gamma w_t -\bar\xB_i^\top \bar\wB_t) \big) \\
    &\ge \ln \big(1+ \exp(-\gamma w_t -c_1\eta)  \big) \\
    &\ge \exp(-\gamma w_t -c_1\eta ) \\
    &= \exp(- \ln \big(e \gamma \eta  \exp(2c_1\eta) (t+1)) \big)\\
    &\ge \frac{1}{ 2 e\gamma \eta \exp(2c_1\eta) (t+1)} \\
    &\ge \frac{1}{c_0\eta \exp(c_0 \eta) t},
\end{align*}
where $c_0 >0$ is independent of $\eta$ and $t$.
\end{proofof}

The next lemma provides a set of examples where the stepsize has to be small for GD to induce a monotonically decreasing risk. 
\begin{lemma}\label[lemma]{lemma:lb:small-stepsize}
Let $\wB_0=0$ and
\[
\bar \xB :=\frac{1}{n}\sum_{i=1}^n \xB_i.
\]
Assume there exist constants $r$ and $q$ such that
\[
\frac{|\{i\in[n]:\xB_i^\top \bar\xB < - r\}|}{n}    \ge q, \quad 
r>0,\ 0<q<1.
\]
Then $L(\wB_1) \le L(\wB_0)$ implies that 
\[
\eta \le \frac{2\ln(2)}{r q}.
\]
\end{lemma}
\begin{proofof}[\Cref{lemma:lb:small-stepsize}]
Since
\(
\wB_0 = 0,\)
we have $L(0) = \ln(2)$ and
\begin{align*}
    \grad L(\wB_0) = -\frac{1}{n}\sum_{i=1}^n \frac{1}{1+\exp(\xB_i^\top \wB_0)}\xB_i = -\frac{1}{2} \bar\xB.
\end{align*}
So we have
\begin{align*}
    \wB_1 = \wB_0  - \eta \grad L(\wB_0) = \frac{\eta}{2} \bar \xB,
\end{align*}
and 
\begin{align*}
    L(\wB_1)
    &= \frac{1}{n} \sum_{i=1}^n \ln \big(1+ \exp(-\xB_i^\top \wB_1)\big) \\
    &=  \frac{1}{n} \sum_{i=1}^n \ln\big( 1+ \exp(-\xB_i^\top \bar\xB \eta / 2)\big).
\end{align*}
If $\eta > 2\ln (2)  /(rq)$, then
\begin{align*}
L(\wB_1) &\ge \frac{1}{n} \sum_{i=1}^n \ln\big( 1+ \exp(-\xB_i^\top \bar\xB \eta / 2)\big)\onebb\big[\xB_i^\top \bar\xB<-r\big] \\ 
&\ge \ln\big( 1+\exp(r \eta / 2) \big) q && \explain{by assumption}\\
&\ge \frac{rq \eta }{2} \\
&\ge \ln(2) = L(\wB_0).
\end{align*}
This proves the original claim by contradiction. 
\end{proofof}

The proof of \Cref{thm:gd:lb} is a combination of \Cref{lemma:lb:small-stepsize,lemma:lb:asymp-rate}.
\begin{proofof}[\Cref{thm:gd:lb}]
We verify that our dataset satisfies the conditions in \Cref{lemma:lb:small-stepsize} with 
\begin{align*}
    r = \frac{1}{10},\quad 
    q = \frac{1}{2}.
\end{align*}
So by \Cref{lemma:lb:small-stepsize} and the monotonic loss, we have
\[\eta \le \frac{2\ln(2)}{rq} \le 40 \ln(2).\]
Then we verify that our dataset satisfies the conditions in \Cref{lemma:lb:asymp-rate} with 
\(
b=0.5.
\)
So by \Cref{lemma:lb:asymp-rate} and the upper bound on $\eta$, we have 
\begin{align*}
    L(\wB_t) \ge \frac{1}{c_0\eta \exp(c_0 \eta) t} \ge \frac{c_1 }{ t},
\end{align*}
where $c_1 >0$ is independent of $\eta$ and $t$ since $\eta$ is upper bounded and $c_0>0$ is independent of $\eta$ and $t$.
\end{proofof}

\section{A Uniform-Convergence Generalization Bound}\label{append:sec:vc-dim}

\begin{proposition}[Uniform convergence]\label[proposition]{prop:gd:linear:generalization}
Suppose \Cref{assump:bounded-separable} holds. Assume that the training samples $(\xB_i, y_i
)_{i=1}^n$ are i.i.d.\ copies of $(\xB, y)$.
Consider a parameter $\hat\wB$ that classifies $(\xB_i, y_i)_{i=1}^n$ correctly, that is, $y_i \xB_i^\top \hat\wB >0$ for every $i$.
Then with probability at least $1-\delta$ over the randomness of $(\xB_i, y_i
)_{i=1}^n$, we have
\begin{align*}
    \Pr(y \xB^\top \hat\wB < 0) \le 4   \frac{d \ln(n+1)+\ln(4/\delta)}{n}.
\end{align*}
In particular, every \cref{eq:gd} output with $L(\wB_t) \le 1/n$ is such a parameter. 
\end{proposition}
\begin{proofof}[\Cref{prop:gd:linear:generalization}]
Consider a hypothesis class of hyperplanes
\begin{align*}
    \Hcal := \big\{ \xB \mapsto \sgn (\xB^\top \wB):  \wB \in \Rbb^d \big\}.
\end{align*}
The Vapnik-Chervonenkis (VC)  dimension of $\Hcal$ is $\vc(\Hcal)=d$~\citep{d-cltem-78}.
Denote the empirical (zero-one) error by 
\[
R_n (\wB) := \frac{1}{n} \sum_{i=1}^n \onebb[ y_i \ne \sgn(\xB_i^\top \wB) ]
= \frac{1}{n} \sum_{i=1}^n \onebb[ y_i \xB_i^\top \wB <0 ],
\]
and the population (zero-one) error by
\[
R(\wB) := \Ebb \onebb[ y \ne \sgn (\xB^\top \wB) ] = \Ebb \onebb[ y\xB^\top \wB <0 ] = 
\Pr(y \xB^\top \wB < 0).
\]
Then by the classical VC-theory for empirical risk minimizer (ERM) \citep[see][Corollary 5.2, for example]{boucheron2005theory}, for every ERM,
\[
\hat \wB \in \arg \min_{\wB} R_n(\wB),
\]
we have the following with probability at least $1-\delta$,
\begin{align*}
    R(\hat \wB) \le R_n(\hat \wB) + 2 \sqrt{R_n(\hat \wB)\frac{2\vc(\Hcal)\ln(n+1) + \ln(4/\delta)}{n}} + 4 \frac{\vc(\Hcal)\ln(n+1) + \ln(4/\delta)}{n}.
\end{align*}
In our case, $R_n(\hat \wB) = 0$, 
so with probability at least $1-\delta$ we have 
\begin{align*}
    R(\hat \wB) \le 4\frac{d\ln(n+1) + \ln(4/\delta)}{n}.
\end{align*}
We complete the proof by noting that $\hat{\wB}$ can be any EMR.
\end{proofof}

\section{Proof of Theorem \ref{thm:sgd:linear}}\label{append:sec:sgd-proof}
Throughout this part, we assume \Cref{assump:sgd:bounded-separable} holds. Without loss of generality, we also assume $y=1$.
For simplicity, we define 
\[
L_\xB (\wB) := \ell(\xB^\top \wB),\quad 
G_\xB(\wB) := \frac{1}{1+\exp(\xB^\top \wB)}.
\]

The following lemma for SGD is an analogy of \Cref{lemma:implicit-bias} for GD.
\begin{lemma}\label[lemma]{lemma:sgd:implicit-bias}
For every $\eta>0$,  $\wB_0$, and  $\uB = \uB_1 + \uB_2$ such that 
\[
\uB_2 = \frac{\eta }{2\gamma} \wB_*,
\]
the following holds for every $t$:
\begin{align*}
    \frac{\|\wB_t - \uB\|^2}{2\eta t} + \frac{1}{t} \sum_{k=0}^{t-1} L_k(\wB_k) \le \frac{1}{t}\sum_{k=0}^{t-1} L_k(\uB_1) + \frac{\|\wB_0 - \uB\|^2}{2\eta t},\quad \as.
\end{align*}
\end{lemma}
\begin{proofof}[\Cref{lemma:sgd:implicit-bias}]
The proof repeats the arguments in \Cref{lemma:implicit-bias}.
By the SGD update rule, we have
\begin{align*}
    \|\wB_{t+1} - \uB \|^2 
    &= \| \wB_t-\uB\|^2 + 2\eta \la \grad L_t(\wB_t) , \uB- \wB_t \ra + \eta^2\big\|\grad L_t(\wB_t) \big\|^2_2 \\
     &= \| \wB_t-\uB\|^2 + 2\eta \la \grad L_t(\wB_t) ,\uB_1 - \wB_t \ra  + \eta\Big( 2\la \grad L_t(\wB_t) ,  \uB_2 \ra + \eta\big\|\grad L_t(\wB_t) \big\|^2_2\Big) .
\end{align*} 
Plugging $\uB_2 = \wB_* \cdot \eta / (2\gamma)$, we can show the second term is non-positive:
\begin{align*}
   &\quad \ 2 \la \grad L_t(\wB_t) ,  \uB_2 \ra + \eta\big\|\grad L_t(\wB_t) \big\|^2_2 \\
   &= 2\ell'(\xB_t^\top\wB_t) \la \xB_t, \uB_2 \ra + \eta\big\| \ell'(\xB_t^\top\wB_t) \xB_t \big\|^2_2 \\
   &\le \eta \ell'(\xB_t^\top\wB_t)   + \eta \big(\ell'(\xB_t^\top\wB_t) \big)^2  && \explain{since \(\uB_2 = \wB_* \cdot \eta /(2 \gamma) \)}\\
   &\le 0, \quad \text{almost surely.} 
\end{align*}
Therefore, we have 
\begin{align*}
    \|\wB_{t+1} - \uB \|^2 
    &\le \| \wB_t-\uB\|^2 + 2\eta \la\grad L_t(\wB_t),  \uB_1-\wB_t \ra  \\
    &\le  \| \wB_t-\uB\|^2 + 2\eta \big( L_t(\uB_1) - L_t( \wB_t) \big),\quad \text{almost surely.} && \explain{by convexity}
\end{align*} 
Summing the above from $0$ to $t-1$ and rearranging, we obtain
\begin{equation*}
   \frac{  \|\wB_t - \uB \|^2}{2\eta} + \sum_{k=0}^{t-1} L_k(\wB_k) 
   \le \sum_{k=0}^{t-1}L_k(\uB_1) + \frac{ \| \wB_0 - \uB\|^2 }{2\eta},\quad \as,
\end{equation*}
dividing both sides by $t$ completes our proof.
\end{proofof}

Similarly, the following lemma for SGD is an analogy of \Cref{lemma:parameter-risk-bound} for GD.
\begin{lemma}[Parameter and risk bounds]\label[lemma]{lemma:sgd:parameter-risk-bound}
Let $\wB_0=0$.
For every $t\ge 0$,  we have
\begin{align*}
    \frac{1}{t}\sum_{k=0}^{t-1} L_k(\wB_k) \le \frac{1 + \ln^2(\gamma^2 \eta t) + \eta^2/4}{ \gamma^2 \eta t},\quad \as,
\end{align*}
and for every $k\le t$,
\begin{align*}
    \|\wB_k\| \le \frac{\sqrt{2} +2\ln(\gamma^2 \eta t) + \eta}{\gamma},\quad \as.
\end{align*}
\end{lemma}
\begin{proofof}[\Cref{lemma:sgd:parameter-risk-bound}]
In \Cref{lemma:sgd:implicit-bias},
choose 
\[
\wB_0 = 0,\quad \uB_1 = \frac{\ln(\gamma^2 \eta t)}{\gamma} \wB_*,
\]
and check that 
\begin{align*}
    \|\uB\| = \frac{\ln(\gamma^2 \eta t) + \eta /2}{\gamma},
\end{align*}
and that for every $\xB$,
\begin{align*}
    L_\xB(\uB_1) &= \ln\big(1+\exp(-\xB^\top \uB_1)\big) \\ 
    &\le \exp(-\xB^\top \uB_1) \\
    &\le \frac{1}{\gamma^2 \eta t},\quad \as.
\end{align*}
Therefore, we have 
\begin{align*}
    \frac{1}{t}\sum_{k=0}^{t-1} L_k(\uB_1) \le  \frac{1}{\gamma^2 \eta t},\quad \as.
\end{align*}
Using \Cref{lemma:sgd:implicit-bias}, we get 
\begin{align*}
    \frac{1}{t} \sum_{k=0}^{t-1}L_k(\wB_k) 
    &\le \frac{1}{t}\sum_{k=0}^{t-1} L_k(\uB_1) + \frac{\|\uB\|^2}{2\eta t} \\
    &\le \frac{1}{\gamma^2 \eta t} + \frac{\ln^2(\gamma^2\eta t) + \eta^2 /4}{\gamma^2 \eta t},\quad \as,
\end{align*}
which verifies the first claim.
For the second claim, we use \Cref{lemma:sgd:implicit-bias} for $s\le t$ and get
\begin{align*}
\text{for every $s\le t$,} \quad
\|\wB_s - \uB\|^2 &\le 2\eta  \sum_{k=0}^{s-1} \ell_k(\uB_1) + \|\uB\|^2 \\
&\le 2\eta s\cdot \frac{1}{\gamma^2 \eta t} + \| \uB\|^2 \\
&\le \frac{2}{\gamma^2} + \| \uB\|^2 ,\quad \as,
\end{align*}
which implies that 
\begin{align*}
\text{for every $s\le t$,} \quad
\| \wB_s \| & \le \|\uB\| + \|\wB_s - \uB \| \\ 
&\le \frac{\sqrt{2}}{\gamma} + 2 \| \uB \| \\ 
&\le \frac{\sqrt{2} + 2\ln (\gamma^2 \eta t) + \eta}{\gamma},\quad \text{almost surely}.
\end{align*}
We complete the proof.
\end{proofof}

The next lemma provides a population risk bound. The main idea is to use martingale concentration tools to convert the ``regret'' bound in \Cref{lemma:sgd:parameter-risk-bound} to a population risk bound.
\begin{lemma}[Population risk bound]\label[lemma]{lemma:sgd:population-risk}
Let $L(\wB) : = \Ebb \ell(\xB^\top \wB)$.
Let $\wB_0 = 0$. 
For every $\eta>0$, with probability at least $1-\delta$ we have 
\begin{align*}
\frac{1}{t}\sum_{k=0}^{t-1}L(\wB_k) 
\le \frac{2+2\ln^2 (\gamma^2 \eta t) + \eta^2 B^4/2}{\gamma^2 \eta  t} +\frac{3 + 2\ln(\gamma^2 \eta t) + \eta}{\gamma}\cdot \frac{18\ln(1/\delta)}{t}.
\end{align*}
\end{lemma}
\begin{proofof}[\Cref{lemma:sgd:population-risk}]
Based on the uniform upper bound on the norm of $(\wB_k)_{k\le t}$ in \Cref{lemma:sgd:parameter-risk-bound}, we can show the following uniform upper bound on risk:
\begin{align*}
\text{for every $k\le t$ and every $\xB$,} \quad
    L_\xB (\wB_k) &= \ln(1+\exp(-\wB_k^\top \xB)) \\
    &\le \ln(1+\exp(\| \wB_k\|)) \\
    &\le \ln (2) + \| \wB_k \| \\ 
    &\le \ln (2) +  \frac{\sqrt{2}+2  \ln (\gamma^2 \eta t) + \eta }{\gamma} \quad \as \\
    &\le M:= \frac{3 +2  \ln (\gamma^2 \eta t) + \eta }{\gamma}.
\end{align*}
Note that $L_\xB(\cdot)$ is non-negative and that $L(\cdot) = \Ebb L_\xB(\cdot)$.
We then have
\begin{align*}
\text{for every $k\le t$ and every $\xB$,} \quad
   | L(\wB_k) - L_{\xB} (\wB_k) | \le M,\quad \as.
\end{align*}
As a direct consequence, we have 
\begin{align*}
    \Ebb \big( L(\wB_k) - L_\xB(\wB_k) \big)^2
    &\le M \cdot \Ebb\big( L(\wB_k) + \ell_\xB(\wB_k) \big) \\
    &= 2M \cdot L(\wB_k),\quad k=0,1,\dots, t.
\end{align*}

Next, we use (one-side) martingale Bernstein inequality (a.k.a.\ Freedman inequality).
With probability at least $1-\delta$, it holds that
\begin{align*}
\sum_{k=0}^{t-1} L(\wB_k) - L_k (\wB_k) 
&\le \sqrt{2 \sum_{k=0}^{k-1} \Ebb \big( L(\wB_k) - L_k(\wB_k) \big)^2 \ln \frac{1}{\delta} }  + \frac{2M}{3} \ln\frac{1}{\delta} \\
&\le \sqrt{ \sum_{k=0}^{t-1} L(\wB_k) \cdot 4M \ln \frac{1}{\delta} }  + \frac{2M}{3} \ln\frac{1}{\delta} \\
&\le \frac{1}{2} \sum_{k=0}^{t-1} L(\wB_t) + 9 M \ln \frac{1}{\delta},
\end{align*}
which implies that 
\begin{align*}
\sum_{k=0}^{t-1}L(\wB_k) 
\le 2 \sum_{k=0}^{t-1} L_k (\wB_k) + 18 M \ln \frac{1}{\delta} .
\end{align*}
Using \Cref{lemma:sgd:parameter-risk-bound} and the definition of $M$, we get
\begin{align*}
\frac{1}{t}\sum_{k=0}^{t-1}L(\wB_k) 
&\le 2\cdot \frac{1+\ln^2(\gamma^2 \eta t) + \eta^2 /4}{\gamma^2 \eta t} + \frac{18M}{t} \ln \frac{1}{\delta}  \\
&\le \frac{2+2\ln^2 (\gamma^2 \eta t) + \eta^2 /2}{\gamma^2 \eta  t} + \frac{18\ln(1/\delta)}{t}\cdot \frac{3 + 2\ln(\gamma^2 \eta t) + \eta }{\gamma},
\end{align*}
which concludes our analysis.
\end{proofof}

The next lemma gives a population zero-one error bound. The proof combines a gradient potential bound and a martingale concentration argument. 
\begin{lemma}[Population error bound]\label[lemma]{lemma:sgd:gradient-bound}
Consider one-pass SGD initialized from $\wB_0=0$.
For every $\eta>0$, with probability at least $1-\delta$, we have
\begin{align*}
\frac{1}{t}  \sum_{k=0}^{t-1} \Pr(y \xB^\top \wB_k)  \le \frac{4\big(\sqrt{2} +2\ln(\gamma^2 \eta t) + \eta \big)}{\gamma^2 \eta t} + \frac{36\ln(1/\delta)}{t}.
\end{align*}
\end{lemma}
\begin{proofof}[\Cref{lemma:sgd:gradient-bound}]
Repeating the perceptron argument as in \Cref{lemma:gradient-bound} , we get 
\begin{align*}
    \frac{1}{t} \sum_{k=0}^{t-1} G_k(\wB_k) \le \frac{\la \wB_t-\wB_0 ,\wB_*\ra}{\gamma \eta t}\le  \frac{\| \wB_t-\wB_0 \|}{\gamma \eta t},\quad \as.
\end{align*}
Using the parameter norm upper bound in \Cref{lemma:sgd:parameter-risk-bound}, we have 
\begin{align*}
    \frac{1}{t} \sum_{k=0}^{t-1} G_k(\wB_k) 
    \le \frac{\| \wB_t -\wB_0 \|}{\gamma \eta t}
    \le \frac{\sqrt{2} +2\ln(\gamma^2 \eta t) + \eta}{\gamma^2 \eta t},\quad \as.
\end{align*}
Note that $0\le G_\xB(\cdot)\le 1$. Applying Freedman's inequality, with probability at least $1-\delta$ we have 
\begin{align*}
    \sum_{k=0}^{t-1} G(\wB_k) - G_k (\wB_k) 
&\le \sqrt{2 \sum_{k=0}^{k-1} \Ebb \big( G(\wB_k) - G_k(\wB_k) \big)^2 \ln \frac{1}{\delta} }  + \frac{2}{3} \ln\frac{1}{\delta} \\
&\le \sqrt{ 2\sum_{k=0}^{t-1}\Ebb \big( G(\wB_k)+ G_k(\wB_k)\big) \ln \frac{1}{\delta} }  + \frac{2}{3} \ln\frac{1}{\delta} \\
&=\sqrt{ \sum_{k=0}^{t-1} G(\wB_k)\cdot 4 \ln \frac{1}{\delta} }  + \frac{2}{3} \ln\frac{1}{\delta} \\
&\le \frac{1}{2} \sum_{k=0}^{t-1} G(\wB_t) + 9  \ln \frac{1}{\delta},
\end{align*}
which implies that 
\begin{align*}
\sum_{k=0}^{t-1}G(\wB_k) 
\le 2 \sum_{k=0}^{t-1} G_k (\wB_k) + 18 \ln \frac{1}{\delta} .
\end{align*}
So with probability at least $1-\delta$ we have 
\begin{align*}
   \frac{1}{t} \sum_{k=0}^{t-1}G(\wB_k)  \le \frac{2\big(\sqrt{2} +2\ln(\gamma^2 \eta t) + \eta \big)}{\gamma^2 \eta t} + \frac{18\ln(1/\delta)}{t}.
\end{align*}
We conclude our proof by noting that zero-one loss is bounded by  $2 G(\cdot)$.
\end{proofof}

The above lemmas imply \Cref{thm:sgd:linear}.
\begin{proofof}[\Cref{thm:sgd:linear}]
It follows from \Cref{lemma:sgd:population-risk,lemma:sgd:gradient-bound}.
\end{proofof}

\section{Proof of Proposition \ref{prop:loss-examples}}\label{append:sec:loss-proof}
\begin{proofof}[\Cref{prop:loss-examples}]
We now verify \Cref{prop:loss-examples}.

\paragraph{Logisitic loss.}
For the logistic loss, we have
\[
\ell(z) := \ln(1+e^{-z}),\quad g(z) := \frac{1}{1+e^z}.
\]
By direct computation, we can verify the first part of \Cref{assump:loss:convex} and \Cref{assump:loss:gradient-bound,assump:loss:exp-tail} for $C_g=1$ and $C_e=2$.
We check the bound on $\rho(\lambda)$ in \Cref{assump:loss:convex} by
\begin{align*}
    \rho(\lambda) 
    &= \min_z \lambda \ell(z) + z^2 \\
    &\le \lambda \ell( \ln( \lambda) ) +  \ln^2( \lambda) && \explain{by setting \(z:=\ln(\lambda)\)}\\
    &\le 1 +  \ln^2( \lambda) .
\end{align*}
We next verify \Cref{assump:loss:local-smoothness} for $C_\beta=e/2$.
Note that $g(z) \le \ell(z) \le C_\beta \ell(z)$.
For the second part of \Cref{assump:loss:local-smoothness}, 
assume $z< x$ without loss of generality (if $x<z$, we apply $g(x) \le g(z)$ and exchange $x$ and $z$ in \Cref{assump:loss:local-smoothness}). 
So we have $x-1\le z < x$ since $|z-x|\le 1$.
By the mean-value form of the Taylor's theorem, we have
\begin{equation}\label{eq:example:local-smooth}
\text{there exists $y\in(z,x)$ such that }   \ell(z) = \ell(x) + \ell'(x)(z-x) + \frac{\ell''(y)}{2}(z-x)^2.
\end{equation}
Note that $\ell''(\cdot) \le g(\cdot)$, $g(\cdot)$ is decreasing, and $|z-x|\le 1$, so we have 
\begin{align*}
    \ell''(y) \le g(y) \le g(z) \le g(x-1) = \frac{e}{e+e^x} \le e\cdot g(x).
\end{align*}
Plugging the above into \eqref{eq:example:local-smooth} verifies \Cref{assump:loss:local-smoothness} for $C_\beta = e/2$.

\paragraph{Flattened exponential loss.}
For the flattened exponential loss with temperature $a>0$, we have 
\begin{align*}
    \ell(z) := \begin{dcases}
        e^{-az} & z>0, \\
        1-az & z \le 0,
    \end{dcases}\qquad
    g(z) := \begin{dcases}
        a e^{-az} & z>0, \\
        a & z \le 0.
    \end{dcases}
\end{align*}
By direct computation, we can verify the first part of  \Cref{assump:loss:convex} and \Cref{assump:loss:gradient-bound,assump:loss:exp-tail} for  $C_g=a$ and $C_e=1/a$.
We check the bound on $\rho(\lambda)$ in \Cref{assump:loss:convex} by
\begin{align*}
    \rho(\lambda) 
    &= \min_z \lambda \ell(z) + z^2 \\
    &\le \lambda \ell\bigg( \frac{\ln(\lambda)}{a} \bigg) +  \frac{\ln^2(\lambda)}{a^2} && \explain{by setting \(z := \ln(\lambda) / a\ge 0\)} \\
    &= 1 +\frac{\ln^2(\lambda)}{a^2}.
\end{align*}
We next verify \Cref{assump:loss:local-smoothness} for $C_\beta=\max\{a, a e^a/2\}$.
The first part of \Cref{assump:loss:local-smoothness} is because
\[g(z) \le a \ell(z) \le C_\beta \ell(z).\]
For the second part of \Cref{assump:loss:local-smoothness}, 
assume $z< x$ without loss of generality.
So we have $x-1\le z < x$.
We discuss two cases.
\begin{itemize}
\item If $z>0$, then $0< z< x$, so $\ell''$ is continuous in $[z,x]$. Then \eqref{eq:example:local-smooth} holds. Using $\ell''(z) = a g(z)$ for $z>0$, $g(\cdot)$ is non-increasing, and $0< z<y< x$, we have
\begin{align*}
    \ell''(y)\le a g(y) \le a g(z) = ae^{-a(z-x)}  g(x)\le 
    a e^a g(x) .
\end{align*}
Plugging the above into \eqref{eq:example:local-smooth} verifies \Cref{assump:loss:local-smoothness}.
\item If $z\le 0$, then $x\le 1$. So 
\begin{align*}
    g(x) \ge g(1) = a e^{-a}.
\end{align*}
Note that $\ell$ is $a^2$-smooth, so we have 
\begin{align*}
    \ell(z) &\le \ell(x) + \ell'(x)(z-x) + \frac{a^2}{2} (z-x)^2 \\
    &\le \ell(x) + \ell'(x)(z-x) + \frac{a e^a}{2} g(x) (z-x)^2,
\end{align*}
which verifies \Cref{assump:loss:local-smoothness}.
\end{itemize}

\paragraph{Flattend polynomial loss.}
For the flattened polynomial loss of degree $a>0$, we have 
\begin{align*}
    \ell(z) := \begin{dcases}
        (1+z)^{-a} & z> 0, \\
        1-a z & z\le 0,
    \end{dcases}\qquad
    g(z) := \begin{dcases}
        a (1+z)^{-(a+1)} & z>0, \\
        a & z \le 0.
    \end{dcases}
\end{align*}
By direct computation, we can verify the first part of \Cref{assump:loss:convex} and \Cref{assump:loss:gradient-bound} for  $C_g=a$.
We check the bound on $\rho(\lambda)$ in \Cref{assump:loss:convex} by
\begin{align*}
    \rho(\lambda) 
    &= \min_z \lambda \ell(z) + z^2 \\
    &\le \lambda \ell\big( \lambda^{1/(a+2)} \big) +  \lambda^{2/(a+2)} && \explain{by setting \(z:=\lambda^{1/(a+2)}\ge 0\)} \\
    &\le 2\lambda^{2/(a+2)}.
\end{align*}
We next verify \Cref{assump:loss:local-smoothness} for $C_\beta=\max\{a, (a+1)2^a)\}$.
The first part of \Cref{assump:loss:local-smoothness} is because
\[g(z) \le a \ell(z) \le C_\beta \ell(z).\]
For the second part of \Cref{assump:loss:local-smoothness}, 
assume $z< x$ without loss of generality.
So we have $x-1\le z < x$.
We discuss two cases.
\begin{itemize}
\item If $z>0$, then $0< z< x$, so $\ell''$ is continuous in $[z,x]$. Then \eqref{eq:example:local-smooth} holds. Using $\ell''(z) \le (a+1) g(z)$ for $z>0$, $g(\cdot)$ is non-increasing, and $0< z<y< x$, we have
\begin{align*}
    \ell''(y)\le (a+1) g(y) \le (a+1) g(z) = (a+1)\bigg(\frac{1+x}{1+z}\bigg)^{a+1}  g(x)\le 
    (a+1)2^{a+1} g(x) .
\end{align*}
Plugging the above into \eqref{eq:example:local-smooth} verifies \Cref{assump:loss:local-smoothness}.
\item If $z\le 0$, then $x\le 1$. So 
\begin{align*}
    g(x) \ge g(1) = a 2^{-(a+1)}.
\end{align*}
Note that $\ell$ is $a(a+1)$-smooth, so we have 
\begin{align*}
    \ell(z) &\le \ell(x) + \ell'(x)(z-x) + \frac{a(a+1)}{2} (z-x)^2 \\
    &\le \ell(x) + \ell'(x)(z-x) + (a+1)2^a g(x) (z-x)^2,
\end{align*}
which verifies \Cref{assump:loss:local-smoothness}.
\end{itemize}

We have verified all examples.
\end{proofof}

The following lemma is useful in our analysis of general losses. 
\begin{lemma}\label[lemma]{lemma:rho}
    Under \Cref{assump:loss:convex}, we have 
    \[
    \ell\big( \sqrt{\rho(\lambda)} \big) \le \frac{\rho(\lambda)}{\lambda}.
    \]
\end{lemma}
\begin{proofof}[\Cref{lemma:rho}]
Recall that 
\[
\rho(\lambda) = \min_{z} \lambda \ell(z) + z^2.
\]
Let $z_*$ be the optimal value. Then 
\begin{align*}
\rho(\lambda) = \lambda \ell(z_*) + z_*^2.
\end{align*}
Since $\ell(\cdot)$ is positive by \Cref{assump:loss:convex}, we have 
\[
z_* \le \sqrt{\rho(\lambda)},\quad 
\ell(z_*) \le \frac{\rho(\lambda)}{\lambda}.
\]
Since $\ell(\cdot)$ is non-increasing by \Cref{assump:loss:convex}, we have
\[
\ell\big(\sqrt{\rho(\lambda)} \big) \le \ell(z_*) \le \frac{\rho(\lambda)}{\lambda},
\]
which completes the proof.
\end{proofof}

\section{Proof of Theorem \ref{thm:ntk}}\label{append:sec:ntk-proof}
We first introduce some notation that will be used extensively in this section.
For a loss function $\ell$ satisfying \Cref{assump:loss:convex,assump:loss:gradient-bound},
define a radius
\begin{equation}\label{eq:ntk:R}
    R:= 6\frac{ \sqrt{\rho(\gamma^2 \eta T)}  + C_a+\sqrt{2\ln(2n/\delta)}  +  \eta  C_g}{\gamma}.
\end{equation}
Define a ball centered at the initialization,
\begin{align*}
    \Bcal := \{ \wB: \|\wB - \wB_0\| \le R\}.
\end{align*}
Define a point-wise linearization error of the network within the ball as
\[
\xi_i(\wB, \vB) :=  f_i( \wB) - f_i( \vB) - \la \grad f_i(\vB), \wB - \vB\ra ,\quad  \wB, \vB \in \Bcal.
\]
Define the maximum linearization error as
\begin{align*}
\xi = \sup_{\wB, \vB \in \Bcal} \sup_{i} \big| \xi_i(\wB, \vB)  \big| 
\end{align*}

Under \Cref{assump:ntk:separable}, we verify that the gradient is bounded by
\begin{equation}\label{eq:ntk:bounded-grad}
    \|\grad f (\xB; \wB)\| = \sqrt{\frac{1}{m} \sum_{s=1}^m a_s^2 \big( \phi'(\xB^\top \wB^{(s)}) \big)^2 \| \xB \|^2} \le 1.
\end{equation}

The proof of \Cref{thm:ntk} breaks down into three parts.
In \Cref{append:sec:ntk-proof:initial}, we control the network properties at the initialization; then in \Cref{append:sec:ntk-proof:linearization}, we bound the linearization error. 
These two parts use standard techniques in the NTK literature.
Results in \Cref{append:sec:ntk-proof:initial,append:sec:ntk-proof:linearization} allow us to show that a good event,
under which the network is well-behaved at initialization and the linearization error is small,
happens with high probability provided there is a large enough network width.
Finally, in \Cref{append:sec:ntk-proof:optimization}, 
we analyze the GD dynamics under general losses assuming that the good event holds.

It is worth noting that our analysis in \Cref{append:sec:ntk-proof:optimization} is decoupled from \Cref{append:sec:ntk-proof:initial,append:sec:ntk-proof:linearization} except assuming the good event holds.
Therefore, our results for GD under general losses in \Cref{append:sec:ntk-proof:optimization} directly apply to a simpler case of training linear models with zero initialization, since the good event automatically holds in this case; we include this consequence explicitly as \Cref{thm:general-loss} in \Cref{append:sec:general-loss}.

\subsection{Initilization Bounds}\label{append:sec:ntk-proof:initial}
The following lemma is from \citep{ji2019polylogarithmic}, showing that the NTK features lead to a large margin under \Cref{assump:ntk:separable}.

\begin{lemma}[NTK feature margin]\label[lemma]{lemma:ntk:init-margin}
Under \Cref{assump:ntk:separable}, 
if $m \ge 50 \ln(n/\delta)/\gamma^2$,
then there exists $\wB_* \in \Rbb^{md}$ with $\|\wB_*\|\le1$ such that with probability at least $1-\delta$, we have
\begin{align*}
\text{for $i=1,\dots,n$},\quad 
    \la y_i \grad f_i(\wB_0), \wB_* \ra \ge \frac{9\gamma}{10} > 0. 
\end{align*}
\end{lemma}
\begin{proofof}[\Cref{lemma:ntk:init-margin}]
This is Lemma 2.3 in \citep{ji2019polylogarithmic}. Choose
\[
\wB_* := 
\begin{pmatrix}
    \frac{a_1}{\sqrt{m}}\chi(\wB_0^{(1)}),
    & \cdots, 
    & \frac{a_m}{\sqrt{m}} \chi(\wB_0^{(m)})
\end{pmatrix}.
\]
By the condition on $\chi$ in \Cref{assump:ntk:separable}, we have $\|\wB_*\|\le 1$.
In addition, note that
\begin{align*}
    \la y_i \grad f_i(\wB_0), \wB_* \ra 
    &= \frac{1}{m} \sum_{s=1}^m y_i \phi'(\xB_i^\top \wB_0^{(s)}) \xB_i^\top \chi(\wB_0^{(s)}).
\end{align*}
By \Cref{assump:ntk:separable}, we have
\begin{align*}
    \Ebb \la y_i \grad f_i(\wB_0), \wB_* \ra  = \Ebb y_i \phi'(\xB_i^\top \wB_0^{(s)}) \xB_i^\top \chi(\wB_0^{(s)}) \ge \gamma.
\end{align*}
By Hoeffding inequality, we have with probability at least $1-\delta$,
\begin{align*}
   \la y_i \grad f_i(\wB_0), \wB_* \ra 
   \ge \Ebb \la y_i \grad f_i(\wB_0), \wB_* \ra
    - \sqrt{ \frac{1}{2{m}} \ln \frac{1}{\delta}
    } \ge \gamma - \sqrt{ \frac{1}{2{m}} \ln \frac{1}{\delta}
    } .
\end{align*}
By a union bound over $i=1,\dots,n$, we have with probability at least $1-\delta$,
\begin{align*}
    \min_{i} \la y_i \grad f_i(\wB_0), \wB_* \ra 
    \ge \gamma - \sqrt{ \frac{1}{2{m}} \ln \frac{n}{\delta}}\ge \frac{9\gamma}{10},
\end{align*}
where the last inequality is because $m\ge 50  \ln(n/\delta) / \gamma^2$
\end{proofof}

The next lemma is also from \citep{ji2019polylogarithmic}, showing that the model output is bounded at initialization.

\begin{lemma}[Initial function value]\label[lemma]{lemma:ntk:init-f}
At initialization, with probability at least $1-\delta$, we have 
\[\max_{i} |f_i(\wB_0)| \le C_a+ \sqrt{2\ln(2n/\delta)}.\]
\end{lemma}
\begin{proofof}[\Cref{lemma:ntk:init-f}]
This is a variant of Lemma 2.4 in \citep{ji2019polylogarithmic}.
Note that $\xB^\top \wB_{0}^{(s)} \sim \Ncal(0, \|\xB\|^2)$, $\|\xB\|\le 1$, and $\phi(\cdot)$ is $1$-Lipschitz, 
so $\phi(\xB^\top \wB_0^{(s)})$ is $1$-subGaussian.
Therefore,
\[
\begin{pmatrix}
    \phi(\xB^\top \wB_0^{(1)} ), &
    \cdots, &
    \phi(\xB^\top \wB_0^{(m)})
\end{pmatrix}^\top
\]
is an $m$-dimensional, $1$-subGaussian random vector.
Note that 
\begin{align*}
    \sqrt{\sum_{s=1}^m \bigg(\frac{a_s}{\sqrt{m}} \bigg)^2 } \le 1, && \explain{by \eqref{eq:ntk:a-condition}}
\end{align*}
so as an inner product of a unit vector and a $1$-subGaussian random vector, 
\begin{align*}
    f(\xB; \wB_0) = \frac{1}{\sqrt{m}}\sum_{s=1}^m a_s \phi(\xB^\top \wB_{0}^{(s)})
\end{align*}
is $1$-subGaussian.
So with probability at least $1-\delta$, we have 
\begin{align*}
    |f(\xB; \wB_0) - \Ebb f(\xB;\wB_0)| \le \sqrt{2 \ln(2/\delta) } \le  \sqrt{2 \ln (2/\delta)}.
\end{align*}
On the other hand, recall that $\xB^\top \wB_{0}^{(s)} \sim \Ncal(0, \|\xB\|^2)$
and that 
\[
Z\sim \Ncal(0, \sigma^2),\quad \Ebb \max\{Z, 0\} = \frac{\sigma}{\sqrt{2\pi}} \le \sigma.
\]
So we have
    \begin{align*}
    \Ebb f(\xB; \wB_0) & = \frac{1}{\sqrt{m}}\sum_{s=1}^m a_s \Ebb \phi(\xB^\top \wB_{0}^{(s)}) \\
    &\le \frac{1}{\sqrt{m}}\sum_{s=1}^m a_s \|\xB\|\\
&\le   C_a . && \explain{since \(\sum_s a_s \le C_a \sqrt{m}\) by \eqref{eq:ntk:a-condition}}
\end{align*}
Applying the triangle inequality and a union bound over $\xB_1,\dots,\xB_n$, we have with probability at least $1-\delta$,
\begin{align*}
    \max_{i } | f_i(\wB_0) | 
    &\le \max_{i} \big(\Ebb f_i(\wB_0) + |f_i(\wB_0) - \Ebb f_i(\wB_0)| \big) \\
    &\le C_a + \sqrt{2\ln({2n}/{\delta} )}.
\end{align*}
This completes the proof.
\end{proofof}

\subsection{Linearization Error Bounds}\label{append:sec:ntk-proof:linearization}

The next lemma is a variant of Lemma 4.1 in \citep{telgarsky2021deep} (which comes from \citep{ji2019polylogarithmic}), connecting the linearization error with the network width $m$. 
The original version in \citep{telgarsky2021deep} only provides a bound on the maximum linearization error. 
Here, we need to use a slightly stronger version that tracks the point-wise linearization error. 

\begin{lemma}[Linearization error]\label[lemma]{lemma:ntk:error-vs-width}
With probability at least $1-\delta$, we have the following for every $\wB, \vB \in \Bcal$:
\begin{align*}
    \sup_{i}|\xi_i(\wB, \vB)| 
&\le \frac{3R^{1/3} + \ln^{1/4}(n/\delta) }{m^{1/6}} \cdot \|\wB - \vB\|,
\end{align*}
and
\begin{align*}
\sup_{i} \|\grad f_i(\wB) - \grad f_i(\vB)\| 
\le    \frac{3R^{1/3} + \ln^{1/4}(n/\delta) }{m^{1/6}}.
\end{align*}
\end{lemma}
\begin{proofof}[\Cref{lemma:ntk:error-vs-width}]
This is a variant of Lemma 4.1 in \citep{telgarsky2021deep}.
The proof is included for completeness. 

Fix $\xB$. Using homogeneity of $\phi$, we have 
\begin{align*}
    f(\xB; \wB) = \frac{1}{\sqrt{m}}\sum_{s=1}^m a_s \phi(\xB^\top \wB^{(s)}) = \frac{1}{\sqrt{m}}\sum_{s=1}^m a_s \ind{\xB^\top \wB^{(s)}>0}\xB^\top \wB^{(s)},
\end{align*}
and 
\begin{align*}
 \lefteqn{   f(\xB; \vB) + \la \grad f(\xB; \vB), \wB - \vB\ra }\\
    &= \frac{1}{\sqrt{m}}\sum_{s=1}^m a_s \phi(\xB^\top \vB^{(s)}) + \frac{1}{\sqrt{m}}\sum_{s=1}^m a_s \ind{\xB^\top \vB^{(s)}>0} \xB^\top \big(   \wB^{(s)} - \vB^{(s)}\big) \\
    &=  \frac{1}{\sqrt{m}}\sum_{s=1}^m a_s \ind{\xB^\top \vB^{(s)}>0}\xB^\top \vB^{(s)} + \frac{1}{\sqrt{m}}\sum_{s=1}^m a_s \ind{\xB^\top \vB^{(s)}>0} \xB^\top \big(   \wB^{(s)} - \vB^{(s)}\big) \\
    &= \frac{1}{\sqrt{m}}\sum_{s=1}^m a_s \ind{\xB^\top \vB^{(s)}>0}\xB^\top \wB^{(s)},
\end{align*}
then we have 
\begin{align*}
\xi_{\xB}(\wB, \vB)
&= f(\xB; \wB) - f(\xB; \vB) - \la \grad f(\xB; \vB), \wB - \vB\ra \\
    &=  \frac{1}{\sqrt{m}}\sum_{s=1}^m a_s \bigg( \ind{\xB^\top \wB^{(s)}>0} - \ind{\xB^\top \vB^{(s)}>0}  \bigg) \xB^\top \wB^{(s)}.
\end{align*}

\paragraph{Control linearization error.}
For $r$ to be determined, define 
\begin{align*}
    \Scal_0 &:= \big\{s\in[m]: |\xB^\top \wB^{(s)}_0| \le r \|\xB\| \big\} \\
    \Scal_1 &:= \big\{s\in[m]: \|\wB^{(s)} - \wB^{(s)}_0\| \ge r \big\} \\
    \Scal_2 &:= \big\{s\in[m]: \|\vB^{(s)} - \wB^{(s)}_0\| \ge r \big\}.
\end{align*}
One can verify that for every $s \notin \Scal_0 \cup \Scal_1 \cup\Scal_2$, 
\begin{align*}
    \sgn\big(\xB^\top \wB^{(s)}\big) =  \sgn\big(\xB^\top \wB^{(s)}_0\big),\quad  \sgn\big(\xB^\top \vB^{(s)}\big) =  \sgn\big(\xB^\top \wB^{(s)}_0\big).
\end{align*}
So for every $s \notin \Scal_0 \cup \Scal_1 \cup\Scal_2$, 
\begin{align*}
    \ind{\xB^\top \wB^{(s)}>0} = \ind{\xB^\top \wB^{(s)}_0>0} =\ind{\xB^\top \vB^{(s)}>0}.
\end{align*}
Then we have 
\begin{align*}
   | \xi_{\xB}(\wB, \vB)|
   &\le \frac{1}{\sqrt{m}}\sum_{s\in [m]} \bigg| \ind{\xB^\top \wB^{(s)}>0} - \ind{\xB^\top \vB^{(s)}>0}  \bigg|\cdot  \big| \xB^\top \wB^{(s)}  \big| \\
   &\le \frac{1}{\sqrt{m}}\sum_{s\in [m]} \bigg| \ind{\xB^\top \wB^{(s)}>0} - \ind{\xB^\top \vB^{(s)}>0}  \bigg|\cdot  \big| \xB^\top \wB^{(s)}  -\xB^\top \vB^{(s)}\big| \\
   &\ \explain{since \(\big| \xB^\top \wB^{(s)}  -\xB^\top \vB^{(s)}\big|\ge\big| \xB^\top \wB^{(s)}  \big| \) if the indicator factor is non-zero} \\
    &\le \frac{1}{\sqrt{m}}\sum_{s\in\Scal_0\cup\Scal_1\cup\Scal_2} \bigg| \ind{\xB^\top \wB^{(s)}>0} - \ind{\xB^\top \vB^{(s)}>0}  \bigg|\cdot  \big| \xB^\top \wB^{(s)}  -\xB^\top \vB^{(s)}\big| \\
    & \ \explain{the indicator factor is zero if \(s\notin\Scal_0\cup\Scal_1\cup\Scal_2\)} \\
    &\le \frac{1}{\sqrt{m}}\sum_{s\in\Scal_0\cup\Scal_1\cup\Scal_2} \big| \xB^\top \wB^{(s)}  -\xB^\top \vB^{(s)}\big| \\
    &\le \frac{1}{\sqrt{m}}\sqrt{|\Scal_0\cup\Scal_1\cup\Scal_2|} \cdot \sqrt{\sum_{s\in\Scal_1\cup\Scal_2\cup\Scal_3} \big| \xB^\top \wB^{(s)}  -\xB^\top \vB^{(s)}\big|^2} \\
    &\le  \frac{1}{\sqrt{m}}\sqrt{|\Scal_0\cup\Scal_1\cup\Scal_2|} \cdot \|\wB - \vB\|.
\end{align*}

\paragraph{Count undesirable indexes.}
Then by Hoeffding inequality, with probability at least $1-\delta$, we have
\begin{align*}
    |\Scal_0| &= \sum_{s=1}^m \ind{ |\xB^\top \wB^{(s)}_0| \le r \|\xB\| } \\
    &\le m \Pr\big\{ |\xB^\top \wB^{(s)}_0| \le r \|\xB\|  \big\} + \sqrt{\frac{m}{2}\ln(1/\delta)}.
\end{align*}
Since $\xB^\top \wB_0^{(s)}/\|\xB\| \sim \Ncal(0, 1)$, we have 
\begin{align*}
    \Pr\big\{ |\xB^\top \wB^{(s)}_0| \le r \|\xB\|  \big\}
    &= \Pr\big\{ |\xB^\top \wB^{(s)}_0|/\|\xB\| \le r   \big\} \\
    &= \frac{1}{\sqrt{2\pi}} \int_{-r}^{r} \exp\big( -z^2 / 2\big)
\dif z \\
&\le \frac{2r}{\sqrt{2\pi}} \le r.
\end{align*}
So with probability at least $1-\delta$,
\begin{align*}
    |\Scal_0| \le m r + \sqrt{\frac{m}{2}\ln(1/\delta)}.
\end{align*}
On the other hand, 
\begin{align*}
    R^2 \ge \sum_{s=1}^m \|\wB^{(s)} - \wB^{(s)}_0\|^2 \ge r^2 |\Scal_1|,
\end{align*}
which implies
\begin{align*}
    |\Scal_1| \le \frac{R^2}{r^2}.
\end{align*}
Similarly, we have
\begin{align*}
    |\Scal_2| \le \frac{R^2}{r^2}.
\end{align*}
So 
\begin{align*}
    |\Scal_0\cup \Scal_1\cup \Scal_2| \le m r + \sqrt{\frac{m}{2}\ln(1/\delta)} + \frac{2R^2}{r^2}.
\end{align*}
Picking $r=R^{2/3} m^{-1/3}$, then we have 
\begin{align*}
    |\Scal_0\cup \Scal_1\cup \Scal_2| &\le m r + \sqrt{\frac{m}{2}\ln(1/\delta)} + \frac{2R^2}{r^2} \\
    &\le m^{2/3} \big( 3 R^{2/3} + \sqrt{\ln(1/\delta)} \big). 
\end{align*}

\paragraph{Union bound.}
Using all the above, we have: with probability at least $1-\delta$, 
\begin{align*}
    |\xi_\xB (\wB, \vB)| &\le  \frac{1}{\sqrt{m}}\sqrt{|\Scal_0\cup\Scal_1\cup\Scal_2|} \cdot \|\wB -\vB\| \\
    &\le \frac{1}{\sqrt{m}}\sqrt{m^{2/3} \big( 3 R^{2/3} + \sqrt{\ln(1/\delta)} \big)} \cdot \| \wB - \vB\| \\
    &\le \frac{3R^{1/3} + \ln^{1/4}(1/\delta) }{m^{1/6}}\cdot \|\wB -\vB\|,\quad \text{for every $\wB,\vB \in \Bcal$}.
\end{align*}
Applying a union bound for the above to hold over $\xB_1,\dots, \xB_n$, we get with probability at least $1-\delta$, 
\begin{align*}
    \sup_{i}|\xi_i (\wB, \vB)| 
    &\le \frac{ 3R^{1/3} + \ln^{1/4}(n/\delta)}{m^{1/6}}\cdot \|\wB -\vB\|,\quad \text{for every $\wB,\vB \in \Bcal$}.
\end{align*}

\paragraph{Gradient bound.}
This part is similar to the previous argument. 
Fix $\xB$, we have 
\begin{align*}
    \|\grad f(\xB; \wB) - \grad f(\xB; \vB)\|^2 
    &= \frac{1}{m}\sum_{s=1}^m \bigg\|\ind{\xB^\top \wB^{(s)}>0}\xB -\ind{\xB^\top \vB^{(s)}>0}\xB \bigg\|^2 \\
    &\le \frac{1}{m}\sum_{s=1}^m \bigg|\ind{\xB^\top \wB^{(s)}>0} -\ind{\xB^\top \vB^{(s)}>0} \bigg|^2 \\
    &\le \frac{1}{m} \cdot |\Scal_0\cup\Scal_1\cup\Scal_2|.
\end{align*}
Applying our bound on $|\Scal_0\cup\Scal_1\cup\Scal_2|$ and a union bound over $\xB$, we get 
\begin{align*}
   \max_{i}\|\grad f_i(\wB) - \grad f_i(\wB)\| 
    \le \frac{1}{\sqrt{m}} \cdot \sqrt{|\Scal_0\cup\Scal_1\cup\Scal_2|}\le \frac{3R^{1/3} + \ln^{1/4}(n/\delta) }{m^{1/6}}
\end{align*}
holds with probability at least $1-\delta$.
\end{proofof}

Using \Cref{lemma:ntk:error-vs-width}, we can show that the linearization error is well-bounded when the network width is sufficiently large. 

\begin{lemma}[Linearization error conditions]\label[lemma]{lemma:ntk:linearization-error}
Recall that
\[
R:= 6 \frac{\sqrt{\rho(\gamma^2 \eta T)}  + C_a+\sqrt{2\ln(2n/\delta)}  +  \eta  C_g}{\gamma}.
\]
Assume that
\[
m \ge  \bigg(\frac{10\big(3R^{1/3} + \ln^{1/4}(n/\delta)\big)}{\gamma}\bigg)^6,
\]
then with probability at least $1-\delta$,
we have the following for every $\wB, \vB \in \Bcal$: 
\[\sup_i |\xi_i(\wB, \vB)| \le\frac{\gamma}{10} \|\wB - \vB\| ,\quad 
\sup_i \|\grad f_i (\wB) -\grad f_i(\vB)\| \le \frac{\gamma}{10}.\] 
\end{lemma}
\begin{proofof}[\Cref{lemma:ntk:linearization-error}]
These follow from our choice of $m$ and \Cref{lemma:ntk:error-vs-width}.
\end{proofof}

\subsection{Optimization Analysis under General Losses}\label{append:sec:ntk-proof:optimization}

\paragraph{The good event and the gradient potential.}
Let $\Ecal$ be the intersection of the events in Lemmas \ref{lemma:ntk:init-margin}, \ref{lemma:ntk:init-f} and \ref{lemma:ntk:linearization-error}.
Then we know $\Ecal$ holds with probability at least $1-3\delta$ if the width $m$ is large enough as specified in \Cref{thm:ntk}.
In what follows, we always operate under the good event $\Ecal$.
We denote the gradient potential by
\begin{align*}
    G(\wB) := \frac{1}{n}\sum_{i=1}^n g \big( y_i f(\xB_i; \wB) \big).
\end{align*}

Our next lemma establishes both a uniform margin bound and a relative margin bound for parameters in a ball.
The relative margin bound is crucial to get the right condition on width $m$ and the ball radius $R$.

\begin{lemma}[Margin in a parameter ball]\label[lemma]{lemma:ntk:uniform-margin}
Under event $\Ecal$,
we have 
\begin{align*}
   \inf_{\wB\in\Bcal} \min_{i}\la y_i \grad f_i (\wB), \wB_*\ra \ge \frac{4\gamma}{5}.
\end{align*}
As a consequence, for 
\[\uB := \vB + \theta \wB_*,\quad \vB \in \Bcal,\]
we have 
\[
\text{for every $i$ and every } \wB \in \Bcal,\quad 
\la y_i \grad f_i(\wB), \uB\ra \ge \frac{\gamma}{10} \big(8\theta - \|\wB - \vB\|\big)  +  y_i f_i(\vB).
\]
\end{lemma}
\begin{proofof}[\Cref{lemma:ntk:uniform-margin}]
The first claim is because
\begin{align*}
    \la y_i \grad f_i (\wB), \wB_*\ra
    &=  \la y_i \grad f_i (\wB_0), \wB_*\ra +  \la y_i \grad f_i (\wB) -\grad f_i(\wB_0), \wB_*\ra \\
    &\ge \frac{9\gamma}{10} - \|\grad f_i (\wB) -\grad f_i(\wB_0)\| && \explain{by \Cref{lemma:ntk:init-margin}} \\ 
    &\ge \frac{4\gamma}{5}. && \explain{by  \Cref{lemma:ntk:linearization-error}}
\end{align*}

To prove the second claim, for $\wB\in\Bcal$, we have 
\begin{align*}
    y_i f_i(\vB) &= y_i \big( f_i(\wB) + \la \grad f_i(\wB), \vB - \wB\ra + \xi_i(\vB, \wB)\big) \\ 
    &= y_i\big(\la \grad f_i(\wB), \vB\ra + \xi_i(\vB, \wB)\big) && \explain{by the homogeneity of \(f\)} \\
    &\le y_i\la \grad f_i(\wB), \vB\ra + \frac{\gamma }{10} \|\wB - \vB\|. && \explain{by \Cref{lemma:ntk:linearization-error}} 
\end{align*}
Therefore, we have 
\begin{align*}
    \la y_i \grad f_i(\wB), \uB\ra 
    &= \la y_i \grad f_i(\wB), \theta \wB_*\ra + \la y_i\grad f_i(\wB), \vB \ra \\
    &\ge \frac{\gamma}{10}(8\theta - \|\wB - \vB\|) + y_i f_i(\vB).
\end{align*}
These complete the proof.
\end{proofof}

The next lemma establishes useful relationships between the gradient potential and the loss. 

\begin{lemma}[Gradient potential properties]\label[lemma]{lemma:ntk:L-G-F}
Consider the loss and the gradient potential.
\begin{enumerate}[leftmargin=*]
\item For $\ell$ satisfying \Cref{assump:loss:local-smoothness}, we have 
\[
G(\wB) \le C_\beta \cdot L(\wB).
\]
\item Under event $\Ecal$, we have
\[ \frac{4\gamma}{5}\cdot G(\wB) \le \|\grad L(\wB)\| \le G(\wB).\]
\end{enumerate}
\end{lemma}
\begin{proofof}[\Cref{lemma:ntk:L-G-F}]
The first claim is by 
\begin{align*}
    G(\wB) &= \frac{1}{n} \sum_{i=1}^n g\big( y_i f_i(\wB) \big) \\
    &\le \frac{C_{\beta}}{n} \sum_{i=1}^n \ell \big( y_i f_i(\wB) \big) && \explain{by \Cref{assump:loss:local-smoothness}} \\
    &= C_\beta L(\wB).
\end{align*}
The upper bound in the second claim is by 
\begin{align*}
    \|\grad L(\wB)\| &=\bigg\|\frac{1}{n} \sum_{i=1}^n \ell'\big( y_i f_i(\wB) \big) y_i \grad f_i (\wB) \bigg\| \\ 
    &\le \frac{1}{n} \sum_{i=1}^n \big| \ell'\big( y_i f_i(\wB) \big) \big|\cdot  \| \grad f_i (\wB) \| \\
    &\le  G(\wB) . && \explain{by \cref{eq:ntk:bounded-grad}}
\end{align*}
The lower bound in the second claim is by 
\begin{align*}
     \|\grad L(\wB)\| &\ge \la  \grad L(\wB), -\wB_*\ra \\ 
     &= - \frac{1}{n} \sum_{i=1}^n \ell'\big( y_i f_i(\wB) \big) \la y_i \grad f_i (\wB), \wB_*\ra \\
     &\ge - \frac{1}{n} \sum_{i=1}^n \ell'\big( y_i f_i(\wB) \big) \cdot \frac{4\gamma}{5} && \explain{by \Cref{lemma:ntk:uniform-margin}} \\ 
     &= \frac{4\gamma}{5} \cdot G(\wB).
\end{align*}
We have completed the proof.
\end{proofof}

In the next lemma, we redo \Cref{lemma:implicit-bias} for nonlinear models with controlled linearization errors and general loss functions.

\begin{lemma}[Split optimization]\label[lemma]{lemma:ntk:implicit-bias}
Consider a loss $\ell$ satisfying \Cref{assump:loss:convex,assump:loss:gradient-bound}.
Assume event $\Ecal$ holds.
Let 
$\uB = \uB_1 + \uB_2$ such that 
\[
\uB_1 = \theta \wB_* + \wB_0,\quad \uB_2 = \frac{2\eta C_g}{\gamma} \wB_*.
\]
For every $t>0$ and $R_t \in (0,  R]$ such that
\[
 \|\wB_k -\wB_0\|\le R_t,\quad k=0,1,\dots, t-1,
\]
we have 
    \begin{align*}
    \frac{\|\wB_{t} - \uB\|^2}{2\eta t} + \frac{1}{t}\sum_{k=0}^{t-1} L(\wB_k) \le 
    \ell\Big(\gamma (8\theta - R_t) / 10 - \big(C_a + \sqrt{2\ln(2n/\delta)}\big)\Big) +  \frac{\|\wB_{0} - \uB\|^2}{2\eta t}.
\end{align*}
\end{lemma}
\begin{proofof}[\Cref{lemma:ntk:implicit-bias}]
For $k<t$, we have
\begin{align*}
    \|\wB_{k+1} - \uB\|^2
    &= \|\wB_{k} - \uB\|^2 + 2\eta \la \grad L(\wB_k), \uB - \wB_k\ra + \eta^2 \|\grad L(\wB_k)\|^2 \\
    &= \|\wB_{k} - \uB\|^2 + 2\eta \la \grad L(\wB_k), \uB_1 - \wB_k\ra  + \eta \big( 2\la \grad L(\wB_k), \uB_2 \ra + \eta \|\grad L(\wB_k)\|^2 \big).
\end{align*}

For the second term, we have 
\begin{align}
\lefteqn{   \la \grad L(\wB_k), \uB_1 - \wB_k\ra} \notag \\
    &= \frac{1}{n} \sum_{i=1}^n \ell'\big(y_i f_i(\wB_t)\big) y_i \la \grad f_i(\wB_k), \uB_1 -\wB_k\ra \notag  \\ 
    &=  \frac{1}{n} \sum_{i=1}^n \ell'\big(y_i f_i(\wB_k)\big) \big( \la y_i \grad f_i(\wB_k),\uB_1\ra - y_if_i(\wB_k)\big) && \explain{by the homogeneity of \(f\)} \notag \\ 
    &\le \frac{1}{n} \sum_{i=1}^n \Big( \ell\big(\la y_i \grad f_i(\wB_k), \uB_1\ra \big) - \ell\big(y_i f_i(\wB_k)\big)\Big)   &&  \explain{by \Cref{assump:loss:convex}} \notag  \\
    &\le \frac{1}{n}\sum_{i=1}^n  \ell\big( \gamma (8\theta-\|\wB_k-\wB_0\|) / 10 + y_i f_i(\wB_0) \big) - L(\wB_t)  && \explain{by \Cref{lemma:ntk:uniform-margin}}\notag\\
     &\le \frac{1}{n}\sum_{i=1}^n  \ell\big( \gamma (8\theta-R_t) / 10 + y_i f_i(\wB_0) \big) - L(\wB_t) . && \explain{since \(\|\wB_k-\wB_0\|\le R_t\)} \label{eq:ntk:convex}
\end{align}
From \Cref{lemma:ntk:init-f} we have 
\[
|y_i f_i(\wB_0)| \le C_a + \sqrt{2\ln(2n/\delta)},
\]
then by the monotinicity of $\ell$ from \Cref{assump:loss:convex}, we have 
\[
\ell\big( \gamma (8\theta-R_t) / 10 + y_i f_i(\wB_0) \big)
\le \ell\Big(\gamma (8\theta - R_t) / 10 - \big(C_a + \sqrt{2\ln(2n/\delta)}\big)\Big).
\]
Therefore, the second term can be bounded by 
\begin{align*}
   \la \grad L(\wB_k), \uB_1 - \wB_k\ra
   \le \ell\Big(\gamma (8\theta - R_t) / 10 - \big(C_a + \sqrt{2\ln(2n/\delta)}\big)\Big) - L(\wB_k).
\end{align*}

For the third term, we have 
\begin{align*}
     \lefteqn{2\la \grad L(\wB_k), \uB_2 \ra + \eta \|\grad L(\wB_k)\|^2} \\ 
     &= \frac{2}{n} \sum_{i=1}^n \ell'\big(y_i f_i(\wB_k)\big) y_i \la \grad f_i(\wB_k), \uB_2\ra  + \eta \bigg\|\frac{1}{n} \sum_{i=1}^n \ell'\big(y_i f_i(\wB_k)\big) y_i \grad f_i(\wB_k) \bigg\|^2\\ 
      &\le \frac{2}{n} \sum_{i=1}^n \ell'\big(y_i f_i(\wB_k)\big) y_i \la \grad f_i(\wB_k), \uB_2\ra  + \eta  \bigg(\frac{1}{n} \sum_{i=1}^n \ell'\big(y_i f_i(\wB_k)\big)\bigg)^2 \qquad \explain{by \cref{eq:ntk:bounded-grad}} \\ 
       &\le \frac{2\|\uB_2\|}{n} \sum_{i=1}^n \ell'\big(y_i f_i(\wB_k)\big) y_i \la \grad f_i(\wB_k), \wB_*\ra  + \eta C_g \cdot G(\wB_k) \qquad \explain{by \Cref{assump:loss:gradient-bound}} \\ 
     &\le - \frac{\gamma \|\uB_2\|}{2}  G(\wB_k)  + \eta C_g \cdot  G(\wB_k) \qquad \explain{by \Cref{lemma:ntk:uniform-margin}} \\
     &\le 0,
\end{align*}
where the last inequality is by the choice of $\uB_2$.

Putting these together, we have for $k<t$
\begin{align*}
     \|\wB_{k+1} - \uB\|^2 \le  \|\wB_{k} - \uB\|^2 + 2\eta \bigg( \ell\Big(\gamma (8\theta - R_t) / 10 - \big(C_a + \sqrt{2\ln(2n/\delta)}\big)\Big) -  L(\wB_k) \bigg).
\end{align*}
Telescoping the sum from $0$ to $t-1$ and rearranging, we get 
\begin{align*}
    \frac{\|\wB_{t} - \uB\|^2}{2\eta t} + \frac{1}{t}\sum_{k=0}^{t-1} L(\wB_k) \le \ell\Big(\gamma (8\theta - R_t) / 10 - \big(C_a + \sqrt{2\ln(2n/\delta)}\big)\Big) +  \frac{\|\wB_{0} - \uB\|^2}{2\eta t},
\end{align*}
which completes the proof.
\end{proofof}

The following lemma combines an induction argument with the arguments in \Cref{lemma:parameter-risk-bound}. 

\begin{lemma}[Risk and parameter bounds in the EoS phase]\label[lemma]{lemma:ntk:parameter-risk-bound}
Consider a loss $\ell$ that satisfies \Cref{assump:loss:convex,assump:loss:gradient-bound}.
Under event $\Ecal$, for every $t\le T$,
\begin{align*}
\frac{1}{t} \sum_{k=0}^{t-1} L(\wB_k) \le 9\frac{\rho(\gamma^2 \eta t) + \big(C_a+\sqrt{2\ln(2n/\delta)}  + \eta C_g \big)^2 }{\gamma^2 \eta t},     
\end{align*}
and
\begin{align*}
    \|\wB_t - \wB_0\| \le 6 \frac{\sqrt{\rho(\gamma^2 \eta t)} + C_a+\sqrt{2\ln(2n/\delta)} + \eta C_g }{\gamma} =: R_t \le  R.
\end{align*}
\end{lemma}
\begin{proofof}[\Cref{lemma:ntk:parameter-risk-bound}]
We prove the claims by induction. 
For $t=0$, the two claims hold.

Now suppose the claims hold for $0,\dots,t-1$, and consider $t$.
By the inductive hypothesis, we have 
\begin{align*}
\text{for}\ k=0,1\dots, t-1,\quad   \|\wB_k-\wB_0\| \le R_k \le R_t := 6 \frac{\sqrt{\rho(\gamma^2 \eta t)} + C_a+\sqrt{2\ln(2n/\delta)} + \eta C_g }{\gamma}.
\end{align*}
So we can invoke \Cref{lemma:ntk:implicit-bias}.
Recall that 
\[
\uB = \uB_1 +\uB_2,\quad 
\uB_1 = \theta \wB_* + \wB_0,\quad 
\uB_2 = \frac{2\eta  C_g}{\gamma} \wB_*.
\]
Let us choose 
\begin{align*}
    \theta = \frac{1}{8} R_t + \frac{5}{4}\frac{ \sqrt{\rho(\gamma^2 \eta t)} + C_a + \sqrt{2\ln(2n/\delta)}}{\gamma} .
\end{align*}
Then we can check that 
\begin{align*}
    \ell\Big(\gamma (8\theta - R_t) / 10 - \big(C_a + \sqrt{2\ln(2n/\delta)}\big)\Big)
    &= \ell\big( \sqrt{\rho(\gamma^2 \eta t)} \big) \\
    &\le \frac{\rho(\gamma^2 \eta t)}{\gamma^2 \eta t}, && \explain{by \Cref{lemma:rho}}
\end{align*}
and that 
\begin{align*}
    \|\uB - \wB_0\| 
    = \theta + \frac{2\eta C_g}{\gamma} 
    =\frac{1}{8} R_t + \frac{5}{4}\frac{ \sqrt{\rho(\gamma^2 \eta t)} + C_a + \sqrt{2\ln(2n/\delta)}}{\gamma}  + \frac{2\eta C_g}{\gamma}.
\end{align*}
Now using \Cref{lemma:ntk:implicit-bias}, we get 
\begin{align*}
    \frac{\|\wB_{t} - \uB\|^2}{2\eta t} + \frac{1}{t}\sum_{k=0}^{t-1} L(\wB_k) 
    &\le  \ell\big( \gamma (8\theta-R_t) / 10 - (C_a + \sqrt{2\ln(2n/\delta)})\big) +  \frac{\|\wB_{0} - \uB\|^2}{2\eta t} \\
    &\le \frac{\rho(\gamma^2 \eta t)}{\gamma^2 \eta t} +\frac{\|\wB_{0} - \uB\|^2}{2\eta t}, 
\end{align*}
which implies that 
\begin{align*}
    \|\wB_t - \wB_0\| 
    &\le \|\wB_t - \uB\| + \|\uB - \wB_0\| \\
    &\le \sqrt{\frac{2\rho(\gamma^2 \eta t)}{\gamma^2}  + \|\wB_0-\uB\|^2}+ \|\uB - \wB_0\| \\
    &\le  \frac{\sqrt{2\rho(\gamma^2 \eta t)}}{\gamma} + 2\|\wB_0-\uB\|\\
    &\le \frac{\sqrt{2\rho(\gamma^2 \eta t)}}{\gamma}
 +    \frac{1}{4} R_t +\frac{5}{2}\frac{ \sqrt{\rho(\gamma^2 \eta t)} + C_a + \sqrt{2\ln(2n/\delta)}}{\gamma}  + \frac{4\eta C_g}{\gamma} \\
&\le  R_t :=  6 \frac{\sqrt{\rho(\gamma^2 \eta t)} + \eta C_g + C_a+\sqrt{2\ln(2n/\delta)}  }{\gamma} ,
\end{align*}
and
\begin{align*}
    \frac{1}{t}\sum_{k=0}^{t-1} L(\wB_k) &\le \frac{\rho(\gamma^2 \eta t)}{\gamma^2 \eta t} + \frac{\|\wB_0 - \uB\|^2}{2\eta t} \\ 
    &\le \frac{\rho(\gamma^2 \eta t)}{\gamma^2 \eta t} + \frac{1}{2\eta t} \bigg( \frac{1}{8} R_t + \frac{5}{4}\frac{ \sqrt{\rho(\gamma^2 \eta t)} + C_a + \sqrt{2\ln(2n/\delta)}}{\gamma}  + \frac{2\eta C_g}{\gamma} \bigg)^2 \\ 
    &\le 9\frac{\rho(\gamma^2 \eta t) + \big( \eta C_g + C_a + \sqrt{2\ln(2n/\delta) }\big)^2}{\gamma^2 \eta t}.
\end{align*}
These verify the claim for $t$.
We have completed the proof.
\end{proofof}

The following lemma is analogous to \Cref{lemma:gradient-bound}.

\begin{lemma}[Gradient potential bound in the EoS phase]\label[lemma]{lemma:ntk:gradient-bound}
Consider a loss $\ell$ satisfying \Cref{assump:loss:convex,assump:loss:gradient-bound}.
Under event $\Ecal$, we have
\begin{align*}
    \frac{1}{t} \sum_{k=0}^{t-1} G(\wB_k)  \le 12 \frac{\sqrt{\rho(\gamma^2 \eta t)} + C_a+\sqrt{2\ln(2n/\delta)} + \eta C_g }{\gamma^2 \eta t},\quad t\le T.
\end{align*}
\end{lemma}
\begin{proofof}[\Cref{lemma:ntk:gradient-bound}]
By \Cref{lemma:ntk:parameter-risk-bound}, we know $\wB_t \in \Bcal$ for $t\le T$.
Then we use the perceptron argument \citep{novikoff1962convergence},
\begin{align*}
    \la \wB_{t+1}, \wB_*\ra &= \la \wB_t, \wB_*\ra  - \eta \la \grad L(\wB_t), \wB_*\ra  \\
    &= \la \wB_t, \wB_*\ra  - \frac{\eta}{n} \sum_{i=1}^n \ell'\big(y_i f_i(\wB_t) \big) \la y_i \grad f_i(\wB_t), \wB_*\ra \\ 
    &\ge  \la \wB_t, \wB_*\ra - \frac{4\gamma \eta}{ 5 n} \sum_{i=1}^n \ell'\big(y_i f_i( \wB_t) \big) && \explain{by \Cref{lemma:ntk:uniform-margin}}\\
    &=  \la \wB_t, \wB_*\ra + \frac{4\gamma \eta}{5} G(\wB_t).
\end{align*}
Telescoping the sum, we have 
\begin{align*}
      \frac{1}{t} \sum_{k=0}^{t-1} G(\wB_k) \le \frac{5 \big( \la \wB_t, \wB_*\ra - \la \wB_0, \wB_*\ra \big) }{4\gamma \eta t}\le \frac{5 \| \wB_t- \wB_0\| }{4\gamma \eta t}.
\end{align*}
Plugging in the parameter bound in \Cref{lemma:ntk:parameter-risk-bound} completes the proof.
\end{proofof}

The following lemma is analogous to \Cref{lemma:stable-phase}.
However, our focus here is a nonlinear predictor that might not be twice differentiable. 
We address this issue using the self-boundedness of the loss in \Cref{assump:loss:local-smoothness} and a slightly longer induction argument. 

\begin{lemma}[Stable phase]\label[lemma]{lemma:ntk:stable-phase}
Consider a loss $\ell$ satisfying \Cref{assump:loss:convex,assump:loss:gradient-bound,assump:loss:local-smoothness}.
Suppose event $\Ecal$ holds.
Suppose there exists $s < T$ such that
\[
L(\wB_s) \le \frac{1}{ 12 C_{\beta}^2 \eta},
\]
then for every $t\in[s, T]$ we have, 
\begin{enumerate}[leftmargin=*]
    \item $G(\wB_t) \le 1/(12 C_{\beta} \eta)$.
    \item $L(\wB_{t+1}) \le L(\wB_t) - \frac{5\eta}{8} \|\grad L(\wB_t)\|^2 \le L(\wB_t)$.
\end{enumerate}
\end{lemma}
\begin{proofof}[\Cref{lemma:ntk:stable-phase}]
Note that $\wB_1,\dots, \wB_T$ all belong to $\Bcal$ by \Cref{lemma:ntk:parameter-risk-bound}.

We first show that Claim 1 implies Claim 2.
Note that 
\begin{align*}
    \|\wB_{t+1} - \wB_t\| &= \eta \|\grad L(\wB_t)\| \\ 
    &\le \eta \cdot G(\wB_t) && \explain{by \Cref{lemma:ntk:L-G-F}} \\ 
    &\le \frac{1}{2 }. && \explain{by Claim 1}
\end{align*}
So we have 
\begin{align*}
    | y_i f_i(\wB_{t+1})- y_if_i( \wB_t) | 
    &= \big|  \la y_i \grad f_i(\wB_t), \wB_{t+1} - \wB_t\ra  + \xi_i(\wB_{t+1}, \wB_t) \big| \\ 
    &\le   \|\wB_{t+1} - \wB_t\|  + \frac{\gamma}{10}\|\wB_{t+1} - \wB_t\| && \explain{by \eqref{eq:ntk:bounded-grad}} \\ 
    &\le \frac{1}{2} + \frac{\gamma}{20} \le 1. && \explain{since \(\gamma\le1\)}
\end{align*}
Denote $z_{i, t} = y_i f_i(\wB_t)$, then $|z_{i, t} - z_{i, t+1}| \le 1$.
By \Cref{assump:loss:local-smoothness} we have
\begin{align*}
\ell(z_{i,t+1}) 
&\le \ell(z_{i,t}) + \ell'(z_{i,t}) (z_{i,t+1} - z_{i,t}) + C_{\beta}  g(z_{i,t})  (z_{i,t+1} - z_{i,t})^2 \\
&\le \ell(z_{i,t}) + \ell'(z_{i,t}) \la y_i \grad f_i(\wB_t), \wB_{t+1} - \wB_t\ra + | \ell'(z_{i,t})| \cdot |\xi_i(\wB_{t+1}, \wB_t)| \\
&\quad +2 C_{\beta}  g(z_{i,t})  \Big( |\la y_i \grad f_i(\wB_t), \wB_{t+1} - \wB_t\ra|^2  + |\xi_i(\wB_{t+1}, \wB_t) |^2 \Big).
\end{align*}
Applying the gradient bound \cref{eq:ntk:bounded-grad} and the local linearization error bound in \Cref{lemma:ntk:linearization-error},
\[
| \xi_{i} (\wB, \vB) | \le \frac{\gamma}{10} \|\wB - \vB\|,\quad \wB, \vB \in \Bcal,
\]
we get 
\begin{align*}
\ell(z_{i,t+1}) 
&\le \ell(z_{i,t}) + \ell'(z_{i,t}) \la y_i \grad f_i(\wB_t), \wB_{t+1} - \wB_t\ra + g(z_{i,t}) \cdot \frac{\gamma}{10}\|\wB_{t+1} - \wB_t\| \\
&\qquad +2 C_{\beta}  g(z_{i,t})  \Big(  \|\wB_{t+1} - \wB_t\|^2  + \frac{\gamma^2}{100}\|\wB_{t+1} - \wB_t\|^2 \Big) \\
&\le \ell(z_{i,t}) + \ell'(z_{i,t}) \la y_i \grad f_i(\wB_t), \wB_{t+1} - \wB_t\ra + g(z_{i,t}) \cdot \frac{\gamma}{10}\|\wB_{t+1} - \wB_t\| \\
&\qquad + 3 C_{\beta}  g(z_{i,t})  \|\wB_{t+1} - \wB_t\|^2,
\end{align*}
where the last inequality is because $\gamma\le 1$.
Taking average over $i=1,\dots,n$, we get
\begin{align*}
  L(\wB_{t+1}) 
  &\le L(\wB_t) + \la \grad L(\wB_t), \wB_{t+1}-\wB_t \ra + G(\wB_t) \cdot \frac{\gamma}{10} \|\wB_{t+1} - \wB_t\| \\
  &\qquad + 3 C_{\beta}  G(\wB_{t+1}) \cdot \|\wB_{t+1} - \wB_t\|^2 \\
  &= L(\wB_t) - \eta \|\grad L(\wB_t)\|^2 +  G(\wB_t) \cdot \frac{\gamma \eta}{10} \|\grad L(\wB_t) \|  \\
  &\qquad + 3 C_{\beta} \eta^2   G(\wB_t) \cdot \|\grad L (\wB_t)\|^2 \\
&\le L(\wB_t) - \eta \|\grad L(\wB_t)\|^2   + \frac{\eta}{8}  \|\grad L(\wB_t)\|^2 \qquad \explain{by \Cref{lemma:ntk:L-G-F}} \\
&\qquad + \frac{\eta}{4} \|\grad L (\wB_t)\|^2 \hspace{30mm} \explain{by \(G(\wB_t) \le 1/( 12 C_{\beta} \eta )\)} \\
&= L(\wB_t) - \frac{5\eta}{8} \|\grad L(\wB_t)\|^2.
\end{align*}
which verifies Claim 2.
We have shown that Claim 1 implies Claim 2.

Next, we prove a stronger version of Claim 1 by induction, that is,
\begin{equation}\label{eq:ntk:F-G-stable}
\text{for every $t\in[s,T] $}, \quad L(\wB_t) \le \frac{1}{12 C_{\beta}^2 \eta},\quad  G(\wB_t) \le \frac{1}{12 C_{\beta} \eta}.
\end{equation}
\begin{itemize}[leftmargin=*]
    \item 
For $t=s$, using our assumption on $L(\wB_s)$ and \Cref{lemma:ntk:L-G-F}, we have 
\begin{equation*}
G(\wB_t) \le C_{\beta}  L(\wB_t) \le \frac{1}{12 C_{\beta}   \eta},
\end{equation*}
which verifies \cref{eq:ntk:F-G-stable} for $s$.

\item 
Now suppose that \cref{eq:ntk:F-G-stable} holds for $s,s+1,\dots,t$.
Since Claim 1 implies Claim 2, we have 
\[
L(\wB_{t+1}) \le L(\wB_t) \le \cdots \le L(\wB_s) \le \frac{1}{12  C_{\beta}^2  \eta}.
\]
From \Cref{lemma:ntk:L-G-F}, we have 
\[
G(\wB_{t+1}) \le C_\beta   L(\wB_{t+1})\le \frac{1}{12 C_{\beta}\eta}.
\]
These together verify \cref{eq:ntk:F-G-stable} for $t+1$ and thus complete our induction.
\end{itemize}
We have proved all claims.
\end{proofof}

The following lemma shows that all data is classified correctly when the training loss is low. 
This lemma 
is only used in the convergence analysis in the stable phase. 

\begin{lemma}\label[lemma]{lemma:ntk:zero-train-error}
Consider a loss $\ell$ satisfying \Cref{assump:loss:convex}.
For every $\wB$ such that 
\[
L(\wB) \le \frac{\ell(0)}{n},
\]
then \(
y_i f_i(\wB) \ge 0\) for $i=1,\dots, n$,
\end{lemma}
\begin{proofof}[\Cref{lemma:ntk:zero-train-error}]
Since $\ell(\cdot) \ge 0$ by \Cref{assump:loss:convex}, we have
\begin{align*}
\frac{1}{n} \ell\big(y_i f_i(\wB_s) \big)
\le L(\wB_s) = \frac{1}{n} \sum_{i=1}^n \ell\big(y_i f_i(\wB_s) \big) \le \frac{\ell(0)}{n},
\end{align*}
which implies 
\[
\ell\big(y_i f_i(\wB) \big) \le \ell(0).
\]
Then the monotonicity of $\ell(\cdot)$ by \Cref{assump:loss:convex} implies
\(y_i f_i(\wB) \ge 0.\)
\end{proofof}

The following lemma is analogous to \Cref{lemma:stable-phase:risk-bound}.
Again, we use an additional induction argument to get a sharp width condition.

\begin{lemma}[Convergence in the stable phase]\label[lemma]{lemma:ntk:stable-phase:risk-bound}
Consider a loss $\ell$ satisfying \Cref{assump:loss:convex,assump:loss:gradient-bound,assump:loss:local-smoothness}.
Suppose event $\Ecal$ holds.
Suppose there exists a time $s < T$ such that 
\begin{align*}
    L(\wB_s) \le \min\bigg\{ \frac{1}{12 C_\beta^2\eta},\ \frac{\ell(0)}{n}\bigg\}.
\end{align*}
Then for every $  0\le t \le T-s$, we have 
\begin{align*}
    L(\wB_{s+t}) \le 16 \frac{\rho(\gamma^2 \eta t)}{\gamma^2 \eta t},\quad 
    \|\wB_{s+t}- \wB_s\| \le 6\frac{\sqrt{\rho(\gamma^2 \eta t)}}{\gamma}.
\end{align*}
\end{lemma}
\begin{proofof}[\Cref{lemma:ntk:stable-phase:risk-bound}]
The first upper bound on $L(\wB_s)$ enables \Cref{lemma:ntk:stable-phase} 
 for $s$ onwards.
Therefore we have for $k\ge 0$,
\begin{align}
    \eta \|\grad L(\wB_{s+k})\|^2 \le 
    \frac{8}{5} \big( L(\wB_{s+k}) -L(\wB_{s+k+1}) \big) 
    \le \frac{8}{5}  L(\wB_{s+k}). \label{eq:ntk:stable}
\end{align}

Next, we use induction to prove the claims. 
For $t=0$, the claims hold.
Consider $t$. By the inductive hypothesis, we have
\begin{align*}
\text{for}\ k=0,\dots, t-1,\quad    \|\wB_{s+k} - \wB_s\| \le 6 \frac{\sqrt{\rho(\gamma^2 \eta k)}}{\gamma} \le 6 \frac{\sqrt{\rho(\gamma^2 \eta t)}}{\gamma} =: R'.
\end{align*}
Choose a comparator centered at $\wB_s$,
\[
\uB := \wB_{s} + \theta \wB_*, \quad \theta := \frac{1}{8}R'+ \frac{5}{4}\frac{ \sqrt{\rho\big(\gamma^2 \eta t\big) }}{ \gamma}.
\]
For $k\le t-1$, we have
 \begin{align*}
     \|\wB_{s+k+1} - \uB\|^2 &= \|\wB_{s+k} - \uB\|^2 + 2\eta \la \grad L(\wB_{s+k}), \uB - \wB_{s+k} \ra + \eta^2 \|\grad L(\wB_{s+k})\|^2 \\
     &\le \|\wB_{s+k} - \uB\|^2 + 2\eta \la \grad L(\wB_{s+k}), \uB - \wB_{s+k} \ra + \frac{8\eta}{5} L(\wB_{s+k}). && \explain{by \eqref{eq:ntk:stable}}
 \end{align*}  
 Repeating the argument in \cref{eq:ntk:convex}, we can bound the second term by
 \begin{align*}
     \la \grad L(\wB_{s+k}), \uB - \wB_{s+k} \ra 
     \le \frac{1}{n}\sum_{i=1}^n  \ell\big( \gamma (8\theta-\|\wB_{s+k}-\wB_s\|) / 10 + y_i f_i(\wB_s) \big)  -  L(\wB_{s+k}).
 \end{align*}
The assumption that $L(\wB_s) \le \ell(0)/n$ allows us to apply \Cref{lemma:ntk:zero-train-error}, so $y_i f_i(\wB_s) \ge 0$ and thus
\begin{align*}
     \gamma (8\theta-\|\wB_{s+k}-\wB_s\|) / 10 + y_i f_i(\wB_s) 
     &\ge  \gamma (8\theta-\|\wB_{s+k}-\wB_s\|) / 10 \\
     &\ge \gamma (8\theta-R') / 10 && \explain{by the inductive hypothesis} \\
     &= \sqrt{\rho(\gamma^2 \eta t)} . && \explain{by the choice of \(\theta\)}
\end{align*}
Using the monotonicity of $\ell$ by \Cref{assump:loss:convex}, we get 
\begin{align*}
   \ell\big( \gamma (8\theta-\|\wB_{s+k}-\wB_s\|) / 10 + y_i f_i(\wB_s) \big) 
   \le \ell\big(\sqrt{\rho(\gamma^2 \eta t)} \big) .
\end{align*}
So we can further control the second term by 
\begin{align*}
     \la \grad L(\wB_{s+k}), \uB - \wB_{s+k} \ra 
     \le  \ell\big( \gamma (8 \theta-R') / 10  \big)  -  L(\wB_{s+k}).
 \end{align*}
 Bringing this back, we get
  \begin{align*}
     \|\wB_{s+k+1} - \uB\|^2 
     &\le \|\wB_{s+k} - \uB\|^2 + 2\eta \Big( \ell\big(\sqrt{\rho(\gamma^2 \eta t)} \big)- L(\wB_t)\Big)   + \frac{8\eta}{5} L(\wB_{s+k}) \\
     &=  \|\wB_t - \uB\|^2 + 2\eta \ell\big(\sqrt{\rho(\gamma^2 \eta t)} \big) -\frac{2\eta }{5} L(\wB_{s+k}) .
 \end{align*}  
Telescoping the sum from $0$ to $t-1$ and rearranging, we get 
\begin{align}
 \frac{5\|\wB_{s+t} - \uB\|^2}{2\eta t} +  \frac{1}{t}\sum_{k=0}^{t-1} L(\wB_{s+k}) 
    &\le 5 \ell\big(\sqrt{\rho(\gamma^2\eta t)}\big) + \frac{5\|\wB_s - \uB\|^2}{2\eta t} \notag \\ 
     &\le 5 \frac{\rho(\gamma^2 \eta t)}{\gamma
     ^2 \eta t}+ \frac{5\|\wB_s - \uB\|^2}{2\eta t}. && \explain{by \Cref{lemma:rho}}. \label{eq:ntk:stable:implicit-bias}
\end{align}
Back to our induction. For case $t$, we have
\begin{align*}
    \|\wB_{s+t}- \wB_s\|
    &\le \|\wB_{s+t} - \uB\| + \|\wB_s- \uB\| \\ 
    &\le \sqrt{ 2\rho(\gamma^2\eta t) / \gamma^2  } + 2\|\wB_s- \uB\| && \explain{by \eqref{eq:ntk:stable:implicit-bias}} \\ 
    &= \frac{\sqrt{2\rho(\gamma^2\eta t)  }}{\gamma} +  \frac{1}{4}R'+ \frac{5}{2}\frac{ \sqrt{\rho\big(\gamma^2 \eta t\big) }}{ \gamma} && \explain{by the choice of \(\uB\)} \\
    &\le R' := 6 \frac{\sqrt{\rho\big(\gamma^2 \eta t\big)} }{ \gamma},
\end{align*}
and
\begin{align*}
  \frac{1}{t}\sum_{k=0}^{t-1} L(\wB_{s+k}) 
    &\le 5 \frac{\rho(\gamma^2 \eta t)}{\gamma
     ^2 \eta t} + \frac{5}{2\eta t} \Bigg( \frac{1}{8} R' + \frac{5}{4} \frac{ \sqrt{\rho\big(\gamma^2 \eta t\big) }}{ \gamma}\Bigg)^2 && \explain{by \eqref{eq:ntk:stable:implicit-bias}} \\
    &= 15\frac{ \rho(\gamma^2 \eta t)}{\gamma^2 \eta t}.
\end{align*}
We complete our induction by using the monotonicity of $L(\wB_t)$ for $t\ge s$ by \Cref{lemma:ntk:stable-phase}.
\end{proofof}

The next lemma shows a (weak) phase transition time bound without using the exponential tail condition of the loss.

\begin{lemma}[Phase transition time]\label[lemma]{lemma:ntk:phase-transition}
Consider a loss $\ell$ satisfying \Cref{assump:loss:convex,assump:loss:gradient-bound}.
Suppose event $\Ecal$ holds.
Define 
\[
\psi(\lambda) = \frac{\lambda}{\rho(\lambda)},\quad \lambda>0.
\]
Then there is $C>0$ as a function of $C_g$, $C_a$, $C_\beta$, $\ell(0)$, $\ln(1/\delta)$ for the following to hold. 
Let 
\begin{align*}
    \tau 
    &:= \frac{1}{\gamma^2 } \max\bigg\{ \frac{\psi^{-1}\big(C(\eta + n) \big)}{\eta},\ C (\eta + n)\eta  \bigg\}.
\end{align*}
If $\tau < T$, then there exists $0\le s \le \tau$ such that
\begin{align*}
    L(\wB_s)  \le \min\bigg\{\frac{1}{12 C_\beta^2 \eta}, \frac{\ell(0)}{n}\bigg\}.
\end{align*}
\end{lemma}
\begin{proofof}[\Cref{lemma:ntk:phase-transition}]
Applying \Cref{lemma:ntk:parameter-risk-bound} with $t=\tau$, we have
\begin{align*}
\frac{1}{\tau} \sum_{k=0}^{\tau-1} L(\wB_k) 
&\le 9 \frac{\rho(\gamma^2 \eta \tau) + \big(C_a+\sqrt{2\ln(2n/\delta)}  + \eta C_g \big)^2 }{\gamma^2 \eta \tau}.
\end{align*}
Choose $\tau$ such that 
\begin{align*}
    \eta \gamma^2 \tau 
    &\ge \max\bigg\{ \psi^{-1}\bigg( 18\bigg(12 C^2_\beta \eta + \frac{n}{\ell(0)} \bigg) \bigg),\\
    &\qquad \qquad \qquad \qquad 18\bigg(12 C^2_\beta \eta + \frac{n}{\ell(0)}\bigg) \big(C_a+\sqrt{2\ln(2n/\delta)}  + \eta C_g \big)^2  \bigg\}.
\end{align*}
It is clear that 
\[
\frac{1 }{\psi(\lambda)} = \frac{\rho(\lambda)}{\lambda} = \min_z \ell(z) + \frac{z^2}{\lambda}
\]
is a decreasing function. 
So for $\eta \gamma^2 \tau$ we have 
\begin{align*}
9\frac{\rho(\gamma^2 \eta \tau)}{\gamma^2 \eta \tau} 
&=  9  \frac{1}{\psi(\eta \gamma^2 \tau) } \\
&\le 9\frac{1}{18\bigg(12 C^2_\beta \eta + {n}/{\ell(0)} \bigg)} \\
&= \frac{1}{2} \cdot \frac{1}{12 C^2_\beta \eta + {n}/{\ell(0)}},
\end{align*}
and 
\begin{align*}
    9 \frac{\big(C_a+\sqrt{2\ln(2n/\delta)}  + \eta C_g \big)^2 }{\gamma^2 \eta \tau}
    &\le \frac{1}{2} \cdot \frac{1}{12 C^2_\beta \eta + {n}/{\ell(0)}}.
\end{align*}
These two inequalities together imply that 
\begin{align*}
    \frac{1}{\tau} \sum_{k=0}^{\tau-1} L(\wB_k) 
&\le 9 \frac{\rho(\gamma^2 \eta \tau) + \big(C_a+\sqrt{2\ln(2n/\delta)}  + \eta C_g \big)^2 }{\gamma^2 \eta \tau} \\
&\le \frac{1}{12 C^2_\beta \eta + {n}/{\ell(0)}} \\
&\le \min\bigg\{\frac{1}{12 C_\beta^2 \eta}, \frac{\ell(0)}{n}\bigg\},
\end{align*}
which implies that there exists $s\le \tau$ for $L(\wB_s)$ satisfies the right hand side bound.
\end{proofof}

The next lemma shows a stronger phase transition bound assuming an exponentially tailed loss.

\begin{lemma}[Phase transition time under exponential tail]\label[lemma]{lemma:ntk:phase-transition-exp}
Consider a loss $\ell$ satisfying \Cref{assump:loss:convex,assump:loss:gradient-bound,assump:loss:exp-tail}.
Suppose event $\Ecal$ holds.
Then there is $C>0$ as a function of $C_e$, $C_g$, $C_a$, $C_\beta$, $\ell(0)$, $\ln(1/\delta)$ for the following to hold. 
Let
\begin{equation*}
    \tau :=  \frac{C}{\gamma^2} \max\big\{\eta,\ n\ln(n) \big\},
\end{equation*}
if $\tau \le T$, then  there exists $0\le s \le \tau$ such that
\begin{align*}
    L(\wB_s)  \le \min\bigg\{\frac{1}{12 C_\beta^2 \eta}, \frac{\ell(0)}{n}\bigg\}.
\end{align*}
\end{lemma}
\begin{proofof}[\Cref{lemma:ntk:phase-transition-exp}]
Under \Cref{assump:loss:exp-tail}, we have,
\begin{align*}
    \ell(z) \le  C_e g(z) = - C_e \ell'(z),\quad z\ge 0,
\end{align*}
which implies 
\begin{align*}
    \frac{\ell'(z)}{\ell(z)} \le -C_e^{-1}, \quad z\ge 0.
\end{align*}
Integrating both sides, we get 
\begin{align*}
    \ln \ell(x) 
    \le \ln \ell(0) + \int_{z =0}^x \frac{\ell'(z)}{\ell(z)} \dif x \le \ln \ell(0)- C_e^{-1} x,\quad x\ge 0,
\end{align*}
which implies 
\begin{align*}
    \ell(x) \le \ell(0) \exp\big(-C_e^{-1} x\big),\quad x\ge 0.
\end{align*}
Using the exponential tail property, we have 
\begin{align}
\rho(\lambda) 
&= \min \lambda\ell(z) + z^2 \notag \\
&\le \lambda \ell\big( C_e \ln(\lambda) \big) + C_e^2 \ln^2(\lambda) \notag \\
&\le \ell(0) + C_e^2 \ln^2(\lambda).\label{eq:ntk:phase-transition:exp-tail}
\end{align}

Applying \Cref{lemma:ntk:gradient-bound} for $\tau$, we have
\begin{align*}
    \frac{1}{\tau} \sum_{k=0}^{\tau-1} G(\wB_k)  &\le 12 \frac{\sqrt{\rho(\gamma^2 \eta \tau)} + C_a+\sqrt{2\ln(2n/\delta)} + \eta C_g }{\gamma^2 \eta \tau} \\ 
    &\le 12 \frac{\sqrt{\ell(0)+ C_e^2 \ln^2(\gamma^2 \eta \tau)} + C_a+\sqrt{2\ln(2n/\delta)} + \eta C_g }{\gamma^2 \eta \tau} && \explain{by \eqref{eq:ntk:phase-transition:exp-tail}} \\
    &\le 12 \frac{ C_e \ln(\gamma^2 \tau)+ \eta (C_g + C_e) +\sqrt{\ell(0)} + C_a+\sqrt{2\ln(2n/\delta)} }{\gamma^2 \eta \tau} \\
    &\le 12\bigg(\frac{C_e}{\eta} \frac{\ln(\gamma^2 \tau)}{\gamma^2 \tau} + \frac{C_g + C_e}{\gamma^2 \tau} + \frac{\sqrt{\ell(0)} + C_a + \sqrt{2\ln(2n/\delta)}}{\eta} \cdot \frac{1}{\gamma^2 \tau} \bigg). 
\end{align*}
So there exists $C>0$ as a function of $C_e$, $C_g$, $C_a$, $C_\beta$, $\ell(0)$, $\ln(1/\delta)$, such that 
\[
\gamma^2 \tau \ge C \max\big\{\eta,\ n\ln(n) \big\}
\]
implies that 
\begin{align*}
 \frac{1}{\tau} \sum_{k=0}^{\tau-1} G(\wB_k)  
        &\le \min\bigg\{\frac{1}{12C_e C_\beta^2 \eta},\; \frac{\ell(0)}{C_e n} \bigg\}.
\end{align*}
So there exists $s \le \tau$ such that 
\[
G(\wB_s) \le  \min\bigg\{\frac{1}{12C_e  C_\beta^2 \eta},\; \frac{\ell(0)}{C_e n}\bigg\}.
\]
The last upper bound ensures that for every $i$,
\begin{align*}
\frac{1}{n} g\big(y_i f_i(\wB_s) \big)
\le G(\wB_s) = \frac{1}{n} \sum_{i=1}^n g\big(y_i f_i(\wB_s) \big) \le \frac{\ell(0)}{C_e n} \le \frac{g(0)}{n},
\end{align*}
where the last inequality is due to \Cref{assump:loss:exp-tail}.
The above implies 
\begin{align*}
y_i f_i(\wB_s) \ge 0
\end{align*}
since $g(\cdot)$ is non-increasing by \Cref{assump:loss:convex}.
So we can apply \Cref{assump:loss:exp-tail} for $y_i f_i(\wB_s)$ and get 
\[
\ell(y_i f_i(\wB_s)) \le C_e g(y_i f_i(\wB_s)).
\]
Taking an average over $i=1,\dots,n$, we get
\[
L(\wB_s) \le C_e G(\wB_s).
\]
We complete the proof by plugging in the upper bound on $G(\wB_s)$.
\end{proofof}

The proof of \Cref{thm:ntk} follows from the above lemmas.
\begin{proofof}[\Cref{thm:ntk}]
It follows from \Cref{lemma:ntk:parameter-risk-bound,lemma:ntk:stable-phase:risk-bound,lemma:ntk:phase-transition,lemma:ntk:phase-transition-exp}.
\end{proofof}





\section{General Losses and Linear Model}\label{append:sec:general-loss}
We restate our \Cref{thm:ntk} for a linear model under general loss functions. 

\begin{theorem}[General losses and linear model]\label{thm:general-loss}
Consider \cref{eq:gd} with stepsize $\eta>0$ for linear classification
\[
L(\wB) := \frac{1}{n}\sum_{i=1}^n \ell(y_i\xB_i^\top \wB ),\quad \wB\in\Rbb^d,
\]
where the dataset $(\xB_i,y_i)_{i=1}^n$ satisfies \Cref{assump:bounded-separable} and the loss function $\ell$ satisfies \Cref{assump:loss:convex,assump:loss:gradient-bound}.
Then we have the following.
\begin{itemize}[leftmargin=*]

\item \textbf{The EoS phase.}~
For every $t > 0$ (and in particular in the EoS phase), we have 
\begin{align*}
\frac{1}{t}\sum_{k=0}^{t-1} L(\wB_k) \le 9\frac{\rho(\gamma^2 \eta t) + \big( \eta C_g \big)^2 }{\gamma^2 \eta t}.
\end{align*}

\item \textbf{The stable phase.}~ 
Assume that the loss $\ell$ also satisfies \Cref{assump:loss:local-smoothness}.
If $s$ is such that 
\begin{equation}\label{eq:linear:stable-phase-criteria}
    L(\wB_s) \le \min\bigg\{ \frac{1}{ 12 C_\beta^2 \eta} ,\; \frac{\ell(0)}{n}\bigg\},
\end{equation}
then \eqref{eq:gd} is in the stable phase, that is, for every $t \in [s, T]$, $L(\wB_t)$ decreases monotonically, and moreover,
\begin{align*}
    L(\wB_t) \le 15 \frac{\rho (\gamma^2 \eta (t-s)) }{\gamma^2 \eta (t-s)}.
\end{align*}
\item \textbf{Phase transition time.}~
There exists a constant $C_1>0$ that only depends on $C_g$, $C_\beta$, and $\ell(0)$ such that the following holds. 
Let 
\begin{equation*}
    \tau := \frac{1}{\gamma^2 } \max\bigg\{ \frac{\psi^{-1}\big(C_1(\eta + n) \big)}{\eta},\ C_1 (\eta + n)\eta  \bigg\},\ \ \text{where}\ \  \psi(\lambda) := \frac{\lambda}{\rho(\lambda)}.
\end{equation*}
If $\tau \le T$, then \cref{eq:linear:stable-phase-criteria} holds for some $s\le \tau$.

\item \textbf{Phase transition time under an exponential tail.}~
Assume that the loss $\ell$ further satisfies  \Cref{assump:loss:exp-tail}, then there exists a constant $C_2>0$ that only depends on $C_e$, $C_g$, $C_\beta$, $\ell(0)$, and $\ln(1/\delta)$ such that the following holds. 
Let
\begin{equation*}
    \tau := \frac{C_2}{\gamma^2} \max\big\{ \eta, \ n \big\}.
\end{equation*}
If $\tau \le T$, then \cref{eq:linear:stable-phase-criteria} holds for some $s\le \tau$. 
\end{itemize}
\end{theorem}
\begin{proofof}[\Cref{thm:general-loss}]
This is proved by reusing results in \Cref{append:sec:ntk-proof:optimization} under a good event where the margin is $\gamma \ge 0.9\gamma$,
the initial predictor is zero,
\[
y_i \xB_i^\top \wB_0 = 0,
\]
the linearization error is zero, and the predictor gradient difference is zero (that is, for $f(\xB; \wB) = \xB^\top \wB$, we have 
$\|\grad f(\xB; \wB) - \grad f(\xB; \vB)\| = 0$).

We remark that different from \Cref{thm:ntk}, here the phase transition time bound under an exponential tail only depends on $n$ instead of $n\ln(n)$.
This difference is because the initial predictor in the NTK case depends on $\ln(n)$, while our initial predictor is zero.
See more details in the proof of \Cref{lemma:ntk:phase-transition-exp}.
\end{proofof}
\end{document}